\documentclass{article}

\usepackage[preprint]{neurips_2025}

\usepackage[utf8]{inputenc} % allow utf-8 input
\usepackage[T1]{fontenc}    % use 8-bit T1 fonts
\usepackage{hyperref}       % hyperlinks
\usepackage{url}            % simple URL typesetting
\usepackage{booktabs}       % professional-quality tables
\usepackage{amsfonts}       % blackboard math symbols
\usepackage{nicefrac}       % compact symbols for 1/2, etc.
\usepackage{microtype}      % microtypography
\usepackage{xcolor}         % colors
\usepackage{graphicx}
\usepackage{amsmath}
\usepackage{wrapfig}
\usepackage{enumitem}
\usepackage{makecell}
\usepackage{float}

\usepackage{array}      % For \newcolumntype and >{\raggedright\arraybackslash} for p, m, b columns
\usepackage{geometry}
\newcolumntype{L}[1]{>{\raggedright\arraybackslash}p{#1}}

\newcommand{\sectionreducemargin}[1]{
\vspace{-3mm} 
\section{#1}
\vspace{-2mm} 
}

\newcommand{\subsectionreducemargin}[1]{
\vspace{-3mm} 
\subsection{#1}
\vspace{-2mm} 
}

\title{ForceVLA: Enhancing VLA Models with a Force-aware MoE for Contact-rich Manipulation}

\author{\normalfont
Jiawen Yu$^{1*}$, Hairuo Liu$^{2,3*}$, Qiaojun Yu$^{4,2\dag}$, Jieji Ren$^2$, Ce Hao$^5$,  Haitong Ding$^6$, \\ Guangyu Huang$^7$, Guofan Huang$^1$, Yan Song$^1$, 
Panpan Cai$^{2,3}$, Wenqiang Zhang$^{1}$, Cewu Lu$^{2,3,8}$\\
$^1$ Fudan University, $^2$Shanghai Jiao Tong University, $^3$Shanghai Innovation Institute,\\ $^4$Shanghai AI Lab,  $^5$ National University of Singapore, $^6$ Shanghai University, \\ $^7$ Xi'an Jiaotong University, $^8$ Noematrix Intelligence\\
$^*$ Equal contribution $^\dag$ Corresponding authors: \texttt{yqjllxs@alumni.sjtu.edu.cn}
}

\begin{document}

\maketitle

% \vspace{-4mm}

\begin{abstract}

Vision-Language-Action (VLA) models have advanced general-purpose robotic manipulation by leveraging pretrained visual and linguistic representations. However, they struggle with contact-rich tasks that require fine-grained control involving force, especially under visual occlusion or dynamic uncertainty. To address these limitations, we propose \textbf{ForceVLA}, a novel end-to-end manipulation framework that treats external force sensing as a first-class modality within VLA systems. ForceVLA introduces \textbf{FVLMoE}, a force-aware Mixture-of-Experts fusion module that dynamically integrates pretrained visual-language embeddings with real-time 6-axis force feedback during action decoding. This enables context-aware routing across modality-specific experts, enhancing the robot's ability to adapt to subtle contact dynamics. We also introduce \textbf{ForceVLA-Data}, a new dataset comprising synchronized vision, proprioception, and force-torque signals across five contact-rich manipulation tasks. ForceVLA improves average task success by 23.2\% over strong $\pi_0$-based baselines, achieving up to 80\% success in tasks such as plug insertion. Our approach highlights the importance of multimodal integration for dexterous manipulation and sets a new benchmark for physically intelligent robotic control. Code and data will be released at \href{https://sites.google.com/view/forcevla2025/}{website}.
\end{abstract}

% sets a new [real-world] benchmark?
\sectionreducemargin{Introduction} \label{Sec: intro}

\begin{figure}
    \centering
    \includegraphics[width=0.9\linewidth]{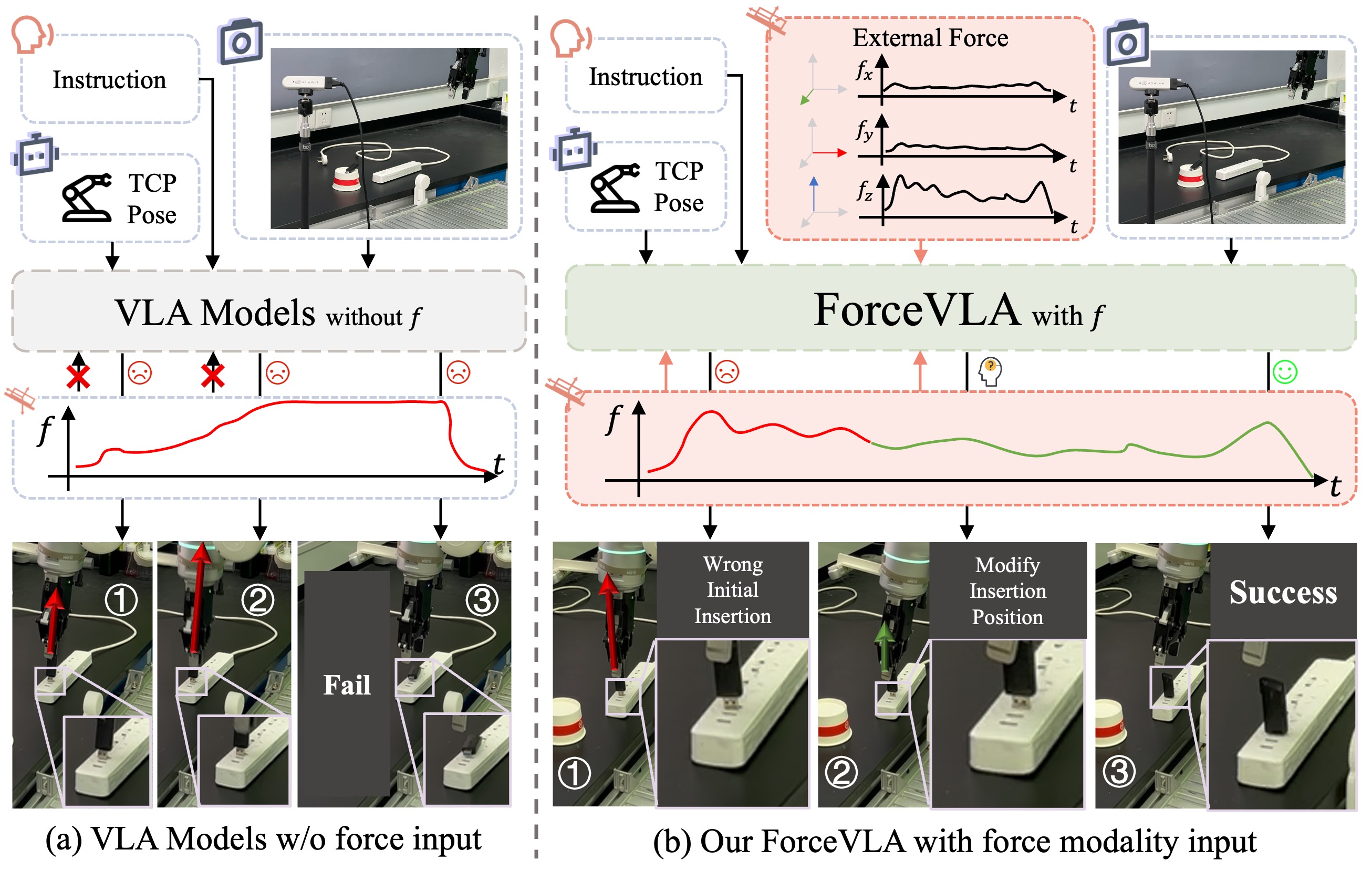}
\vspace{-4mm}
    \caption{Comparison between ForceVLA and baselines without force input. Without force feedback, the policy fails to correct pose errors and completes insertion incorrectly. In contrast, ForceVLA leverages external force signals to adjust insertion strategies dynamically, leading to successful execution despite initial misalignment.}
    \label{Fig: teaser}
\vspace{-4mm}
\end{figure}

Robotic learning has advanced rapidly with the rise of embodied AI, driven by large-scale manipulation datasets and the emergence of foundation models~\cite{droid, openx-embodiment, Bridgedatav2}. These models exhibit strong adaptability, enabling rapid generalization to novel tasks with minimal supervision~\cite{rt1, team2024octo, brohan2023rt}. In parallel, Vision-Language Models (VLMs) have achieved remarkable generalization through large-scale multimodal alignment~\cite{radford2021learning, zhuang2025math}, learning transferable representations that support a wide range of downstream tasks.

Building on this progress, OpenVLA~\cite{openvla} introduced Vision-Language-Action (VLA) models to bridge perception and control for real-world robotic manipulation. By leveraging VLM-based encoders, these models demonstrate strong performance in semantic grounding, language following, and zero-shot generalization. $\pi_0$~\cite{pi_0} further enhances this framework using stronger VLM backbones~\cite{paligemma} and flow-based action generation, showing that pretrained multimodal VLA models can acquire robust physical-world priors and can be fine-tuned efficiently with only a few demonstrations.
 
% However, contact-rich manipulation involves more than just semantic grounding and spatial planning—it is fundamentally governed by interaction forces~\cite{contact1, noseworthy2025forge}. Existing VLA models, which rely heavily on visual and linguistic cues, often overlook this crucial modality, despite its essential role in contact-rich physical manipulation tasks. In contrast, humans naturally integrate multimodal feedback—including interaction forces—when executing such tasks, continuously adapting their manipulation strategies based on tactile and proprioceptive signals~\cite{kim2015multimodal}. Consequently, existing VLA models often struggle with precision tasks such as insertion, tool use, and assembly, particularly under occlusion or visual degradation, where vision alone fails to capture the underlying physical dynamics. These limitations frequently result in brittle behaviors, suboptimal motion planning, or even complete task failure. Moreover, the force requirements in contact-rich manipulation tasks vary dynamically across different phases. For example, grasping may necessitate delicate contact, whereas insertion requires precisely controlled force to overcome friction and determine whether the task has been completed. Surface interactions, such as wiping, demand sustained and compliant pressure. Yet, existing methods lack the mechanisms to perceive and adapt to these phase-dependent variations in force, thereby limiting their capacity to reason about physical interactions over time.

However, contact-rich manipulation demands more than semantic grounding and spatial planning—it is fundamentally driven by interaction forces~\cite{contact1, forge}. Existing VLA models rely heavily on visual and linguistic cues, often overlooking force sensing, a modality critical for precise physical interaction. In contrast, humans naturally integrate tactile and proprioceptive feedback to adapt their manipulation strategies~\cite{kim2015multimodal}. As a result, VLA models frequently struggle with tasks such as insertion, tool use, or assembly—especially under occlusion or poor visual conditions—leading to brittle behavior or task failure. Moreover, force requirements evolve across different task phases: delicate grasping, controlled insertion, and compliant surface contact—each requiring distinct forms of force modulation. Current methods lack mechanisms to perceive and adapt to these dynamic variations, limiting their ability to reason over time about physical interactions.

To address these limitations, we introduce \textbf{ForceVLA}, a novel framework that augments VLA models with a force-aware Mixture-of-Experts (MoE) module, enabling effective reasoning and context-sensitive, force-informed action generation in contact-rich manipulation tasks, as illustrated in Figure~\ref{Fig: teaser}. \textbf{ForceVLA} is grounded in the key insight that 6D external force sensed at the robot’s end-effector, should be treated as a first-class modality and formally integrated into the action expert module to enable phase-aware action generation based on force feedback during physical interaction. To realize this integration, ForceVLA incorporates a force-aware MoE module, named \textbf{FVLMoE}, designed to perform modality- and phase-aware fusion of visual-linguistic representations with real-time force feedback from embodied interaction during action planning. Through a gating mechanism, \textbf{FVLMoE} computes dynamic routing weights over expert subnetworks, each specialized for different modalities across task execution phases.  By adaptively activating these experts based on high-level task instructions and low-level interaction feedback, ForceVLA captures subtle yet critical, phase-dependent variations during physical interaction and generates precise, phase-aligned, and force-aware action chunking. Our main contributions are:

\vspace{-2mm}
\begin{itemize}[leftmargin=15pt]
\item We present a novel framework that integrates force, vision, language, and action for improved precision and stability on contact-rich manipulation tasks. Key to our approach is a force-aware Mixture-of-Experts-based fusion module, which enables dynamic processing and deep integration of force, visual, and language features during action generation, significantly enhancing physical interaction capabilities in VLA systems.
\item  We build a complete data collection pipeline—including teleoperation tools, data converters, and a new dataset—specifically for contact-rich manipulation, and commit to open-sourcing all resources for community use.
\item Through experiments on five challenging tasks, ForceVLA achieves up to 80\% task success and improves average performance by 23.2\% over baselines, demonstrating strong generalization to novel objects, occlusions, and physical perturbations.
\end{itemize}

\sectionreducemargin{Related Works} \label{Sec: related works}

\textbf{Robotic VLA domain.}
Recent research in Vision-Language-Action (VLA) models has focused on leveraging large-scale multimodal pretraining to generalize robotic policies across tasks and embodiments~\cite{rt1,brohan2023rt,openvla,openvla-oft,vlas,robomamba,vima,gr1,interleave}. These models typically map visual and language inputs to low-level control signals via end-to-end learning. Flow-based architectures such as $\pi_0$~\cite{pi_0,pi0.5} integrate pretrained vision-language encoders with fast action decoders to achieve high-frequency outputs. Other works incorporate reasoning mechanisms~\cite{cot-vla,dexvla,robodual,fastpi}, action space compression, or 3D point cloud inputs~\cite{pointvla} to improve instruction grounding and task execution. Diffusion-based models~\cite{team2024octo,dp,prediction,chatvla,hybridvla} introduce stochastic generation for diverse, long-horizon behaviors, though they often incur high training and inference costs. Despite these advances, most VLA approaches remain limited to vision and language inputs, making them less effective in contact-rich or occluded manipulation scenarios where tactile feedback is critical.

\textbf{Contact-rich manipulation domain.}
Traditional vision-only methods struggle with dynamic interactions requiring fine-grained feedback. To address this, recent works integrate force sensing~\cite{tacdiffusion,adaptiveCP,forcemimic,force-aware}, enabling improved motion stability and accuracy. Xie et al.~\cite{towards,forge} provide foundational studies on the role of force feedback in robotic control. Tactile sensing has also emerged as a powerful modality: TLA~\cite{tla} and Tac-Man~\cite{tac-man} demonstrate enhanced performance in fine manipulation and articulation tasks. Multimodal fusion methods~\cite{impact,should} show promise in complex environments, though current approaches are often limited to static modality fusion and lack dynamic routing or unified modeling frameworks. Furthermore, few evaluate cross-task generalization in real-world contact-rich settings.

% \textbf{MoE architecture-related work.}
% %
% Sparsely activated Mixture-of-Experts (MoE) architectures have shown great potential to improve model capacity and computational efficiency. Foundational works~\cite{Outrageously,switch,gshard,glam,sparsemoe} 
% have facilitated the application of MoE in large-scale language models, achieving significant parameter expansion and improved training scalability. Zoph et al.~\cite{st-moe} further addresses stability and transferability issues during MoE training, improving adaptability across multitask settings. For multimodal learning, Mustafa et al.~\cite{limoe} incorporate sparse expert layers to jointly learn from image and text modalities, enhancing the performance of multimodal contrastive learning. In the domain of robot control, recent works~\cite{more,chatvla} introduce MoE architectures into vision-language-action (VLA) models, enabling improved policy generalization and environmental adaptability. However, these methods have yet to incorporate force sensing as an explicit modality, nor have they realized end-to-end modeling of dynamic multimodal interactions involving fine-grained physical contact.

\textbf{MoE architecture-related work.}
Mixture-of-Experts (MoE) architectures improve model scalability and efficiency by activating sparse expert subnetworks~\cite{Outrageously,switch,gshard,glam,sparsemoe}. Follow-up work~\cite{st-moe} improves MoE training stability and task transferability. In the multimodal domain, LIMOE~\cite{limoe} integrates sparse expert layers for joint vision-language learning. Recent applications in robotics~\cite{more,chatvla} adopt MoE layers within VLA models to enhance policy generalization and adaptability. However, these methods largely omit explicit modeling of the force/tactile modalities, and lack mechanisms for dynamically routing across multimodal signals in contact-intensive tasks.

% \textbf{Positioning Our Work:}
% Pi0 serves as our architectural foundation but exhibits suboptimal performance on contact-rich tasks. Rdp (Robotic Dexterous Policies, or specify the actual paper) also introduces force modality but lacks the generalization of VLAs. Our work effectively fuses force and VLA modalities, achieving both the fine-grained manipulation capabilities of task-specific models and the robust generalization of general-purpose VLA models.

\sectionreducemargin{Preliminary} \label{Sec: prelim}

\begin{wrapfigure}{R}{0.5\textwidth}
  \centering
  \includegraphics[width=0.48\textwidth]{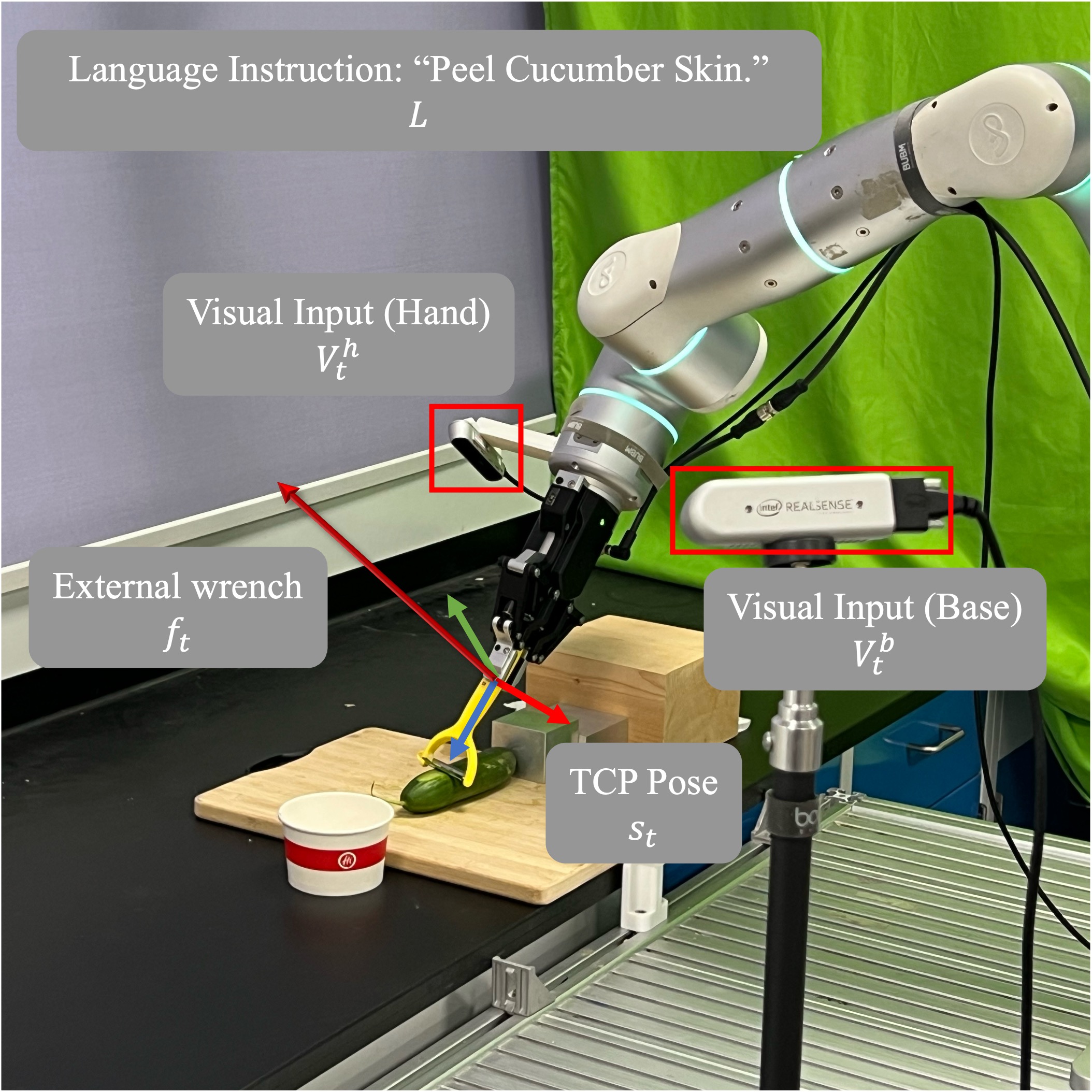}
  \caption{Robot manipulation tasks setting.}
  \label{Fig: illustration of prelim}
\vspace{-3mm}
\end{wrapfigure}
\textbf{Problem Formulation.} 
Figure~\ref{Fig: illustration of prelim} shows the setting of robot manipulation tasks.
The robot's observation at timestep $t$ consists of base and hand visual inputs $V_t^b$ and $V_t^h$, the proprioceptive state $s_t \in \mathbb{R}^7$, and external force-torque readings $f_t \in \mathbb{R}^6$, which are collectively denoted as $O_t = \{V_t^b, V_t^h, s_t, f_t\}$. 
Given a language instruction $L$, the objective is to learn an end-to-end policy $\pi(A_t | O_t, L)$ that outputs low-level, executable action chunk $A_t=\{a_t, a_{t+1},...,a_{t+H-1}\}$\cite{pi_0} maximizing the likelihood of completing the contact-rich task, where $s_t$ is a vector of TCP pose concatenated with gripper width. TCP position is represented by Cartesian coordinates $(x,y,z)$ and orientation is represented by Euler angles $(\alpha,\beta,\gamma)$. $f_t$ is the estimated external wrench applied on TCP and expressed in world frame, which consists of $\mathbb{R}^3$ force and $\mathbb{R}^3$ moment: $f_t=\{f_{tx},f_{ty},f_{tz},m_{tx},m_{ty},m_{tz}\}$. 

\textbf{MoE Architecture.} 
We select Mixture-of-Experts (MoE)\cite{sparsemoe, gshard} as our fusion layer. The core idea is to distribute different modalities to a larger set of smaller, specialized ``expert'' subnetworks, only a fraction of which are activated for any given input token.
An MoE layer typically comprises a set of $N$ \textbf{expert networks}, denoted as $\{E_i\}_{i=1}^N$ and a \textbf{gating network} (also referred to as a router), denoted as $G$. This network takes an input token $x$ and dynamically determines which of the $N$ experts should process it. In prevalent sparse MoE implementations, for an input token $x$, the gating network $G(x)$ produces scores or logits that are used to select a small subset of $k$ experts (typically $k=1$ or $k=2$, where $k \ll N$) from the total pool of $N$ experts. The input token $x$ is then routed only to these $k$ active experts. The outputs of these active experts, $E_i(x)$, are subsequently aggregated, commonly through a weighted sum where the weights $g_i(x)$ are also derived from the gating network. The final output $y(x)$ of the MoE layer can be expressed as:
$y(x) = \sum_{i \in \text{TopK}(G(x))} g_i(x) E_i(x)$, where $\text{TopK}(G(x))$ denotes the set of indices of the top-$k$ experts selected by the gating network for input $x$.
% %
% This paradigm not only enhances model scalability but also promotes the \textbf{specialization of experts} to distinct data sub-domains, features, or functionalities. The ability of MoE to increase capacity with sub-linear growth in computational cost has made it a key technique in scaling large language and multimodal models.

\sectionreducemargin{ForceVLA} \label{Sec: method}
\vspace{2mm}

\subsectionreducemargin{Overview of ForceVLA}

\begin{figure}
    \centering
    \includegraphics[width=0.95\linewidth]{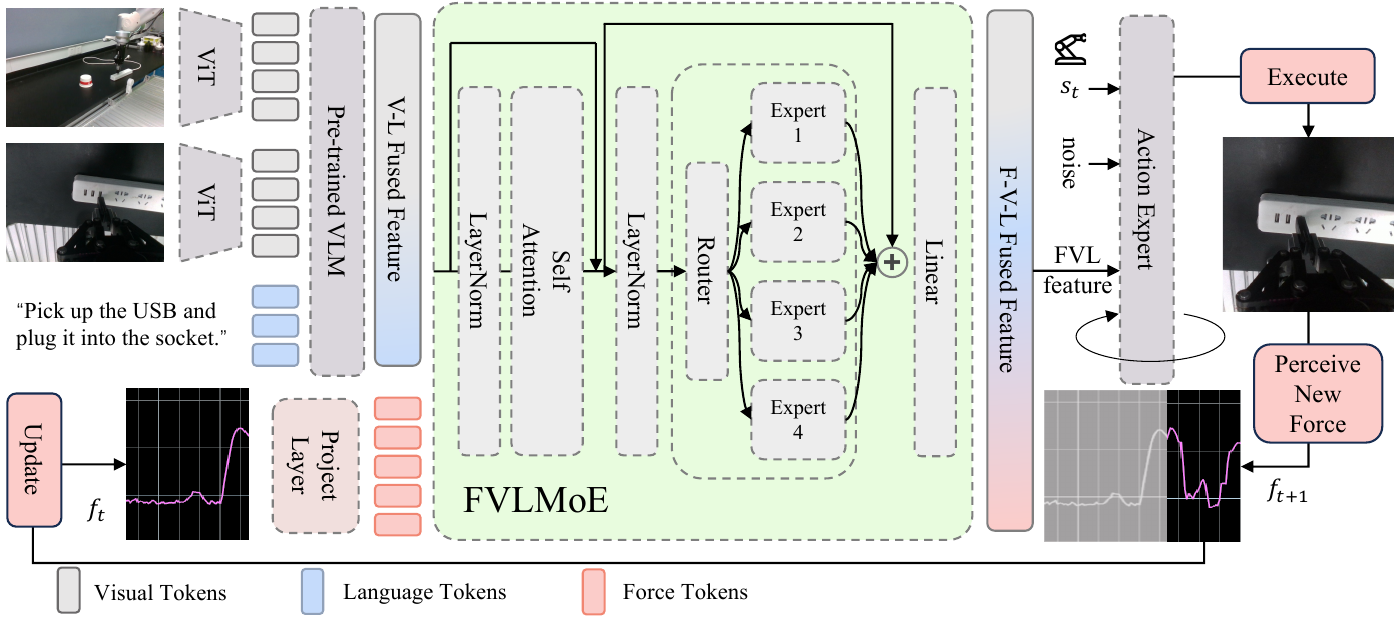}
    \caption{Overview of our ForceVLA model. Visual and language inputs are processed by a pre-trained VLM to form contextual embeddings. External force signals are projected and fused with VLM outputs via the FVLMoE module. The resulting multimodal features guide a flow-based action head to generate contact-aware robot actions.}
    \label{Fig: pipeline}
\vspace{-6mm}
\end{figure}

ForceVLA is an end-to-end multimodal robotic policy designed for contact-rich manipulation. Its pipeline is illustrated in Figure~\ref{Fig: pipeline}. Building upon the $\pi_0$ framework~\cite{pi_0}, it integrates vision, language, proprioception, and 6-axis force feedback to generate actions through a conditional flow matching model~\cite{rectifiedflow,flowmatching}. Visual inputs from multiple RGB cameras and task instructions are encoded by a SigLIP-based~\cite{siglip} vision-language model (based on PaliGemma~\cite{paligemma}) into contextual embeddings. These embeddings, combined with proprioceptive and force cues, condition an iterative denoising process that predicts the action trajectory.

\textbf{FVLMoE} is the core module enabling effective force integration. Force readings are linearly projected into dedicated tokens and fused with vision-language embeddings via a Mixture-of-Experts (MoE) module. Inspired by MoE’s strength in multi-task and modality-specific learning~\cite{moemultitask,limoe}, FVLMoE adaptively routes and processes multimodal inputs. Its output serves as a rich guidance signal for the flow model, allowing ForceVLA to handle subtle contact dynamics and visually ambiguous scenarios with greater precision and robustness.

\subsectionreducemargin{FVLMoE Architecture}
The FVLMoE module is specifically designed for the fusion of multimodal sensory information. Its design enables the model to integrate rich contextual understanding from vision and language with the immediate, fine-grained dynamics captured by force-torque sensors. This fusion is critical for enabling robust and adaptive behavior in contact-rich manipulation tasks. The architecture and operation of the FVLMoE can be detailed in the following stages:

\textbf{Input Mapping for Multiple Modalities.}
How to determine the optimal stage and method for a novel force modality incorporation poses a significant design challenge. Following extensive experimentation, we established an approach where the force modality is introduced after the primary VLM has processed visual and linguistic inputs. Specifically, force features are fed as distinct inputs into the FVLMoE module, positioning it to function akin to a higher-level cortical association area responsible for integrating the VLM's pre-trained visual-linguistic representations with the newly introduced force tokens. This strategy contrasts with introducing force prior to, or concurrently with, the VLM's initial fusion of visual and language modalities. The empirical justification for this architectural decision is elaborated in the Ablation Studies section (Section~\ref{Sec: ablation}).

The FVLMoE module, in line with this design choice, ingests a sequence of token embeddings $E_{in}$ formed by the concatenation of visual-linguistic features and a dedicated force token. The VL features, denoted as $E_{VL} \in \mathbb{R}^{N_{VL} \times D_{model}}$, are outputs from the primary Vision-Language Model, encapsulating contextual understanding derived from processed image streams and textual instructions. Concurrently, the raw 6-axis force-torque sensor data, $f_{raw} \in \mathbb{R}^{6}$, is transformed by a linear projection $\phi_F$ into a force token embedding $E_F = \phi_F (f_{raw}) \in \mathbb{R}^{D_{{model}}}$. The final input to the FVLMoE is thus the concatenated sequence $E_{in} = [E_{VL} ; E_F] \in \mathbb{R}^{(N_{VL}+1) \times D_{model}}$, where the force token is appended to the visual-linguistic context for subsequent joint processing within the MoE architecture.

\textbf{Multimodal Routing and Fusion Computation.}
Once the combined multimodal sequence $E_{{in}} \in \mathbb{R}^{(N_{{VL}}+1) \times D_{{model}}}$ is formed, it undergoes hierarchical processing within the FVLMoE module. $E_{{in}}$ is passed through an encoder layer for shared refinement to facilitate holistic interaction among all constituent force, visual, and language tokens. This layer is composed of a multi-head self-attention mechanism with $N_{{heads}}$ attention heads and a subsequent FFN, yielding $E_{{enc}} \in \mathbb{R}^{D_{{model}}}$. Subsequently, $E_{{enc}}$ is channeled into a sparse Mixture-of-Experts layer. This layer employs $E=4$ distinct expert networks, each realized as an independent MLP. A dynamic gating network determines the routing, selecting the most appropriate single expert (top $k=1$) for each token in $E_{{enc}}$ based on learned dispatch weights. The output from the MoE computation is then integrated back with the input to the MoE layer via a residual connection, yielding $E_{{fused}}$. The resulting sequence of fused multimodal features is passed through a final linear projection layer to match the dimensionality of the action expert.

\textbf{Injecting Fused Features into the Action Flow Head.}
The sequence of fused multimodal features produced by the FVLMoE module serves as a guidance signal for the action generation process, which is formulated as a flow-based denoising model. This guidance is materialized by first extracting a specific sub-sequence, $G_{\text{FVLMoE}} \in \mathbb{R}^{H_{\text{action}} \times D_{\text{a}}}$, comprising the final $H_{\text{action}}$ tokens from $E_{\text{FVLMoE}}$; these tokens encapsulate the most pertinent fused guidance for each step in the $H_{\text{action}}$-length action plan. $G_{\text{FVLMoE}}$ is then combined via element-wise addition with $S_{\text{suffix}} \in \mathbb{R}^{H_{\text{action}} \times D_{\text{a}}}$ obtained from the primary VLM's processing of the current proprioceptive robot state $s_t \in \mathbb{R}^{D_s}$ and the noisy action trajectory $a_{t}^{\tau} \in \mathbb{R}^{H_{\text{action}} \times D_a}$ at denoising step $\tau$, where $D_s$ and $D_a$ are the dimensionalities of the state and action spaces, respectively. This additive injection mechanism ensures that the rich, contact-aware contextual understanding developed by the FVLMoE directly modulates and refines the generated action sequence at each step of the predicted trajectory.

\subsectionreducemargin{Datasets}

To train ForceVLA, we curated a new dataset specifically focused on contact-rich manipulation tasks, emphasizing the synchronized capture of visual, proprioceptive, and force-torque data. Existing datasets often lack the comprehensive force interactions or the diversity of contact-driven scenarios necessary to develop robust force-aware policies.

% \textbf{Data Collection Platform.}
Our data collection was performed using a Flexiv Rizon 7-DOF robotic arm equipped with a Dahuan adaptive gripper. Visual data was captured from two RGB-D cameras: one static third-person view (RealSense D435 at 1280x720, 30 FPS) and one wrist-mounted camera (RealSense D415 at 640x480, 30 FPS) providing egocentric perspectives. Data was collected via human teleoperation using a Quest3 VR interface with custom mappings to robot end-effector control. Five expert operators performed a total of 5 distinct contact-rich tasks: \textit{bottle pumping, plug insertion, USB drive insertion, whiteboard wiping,} and \textit{cucumber peeling}, as described in Section~\ref{sec:evaluation_tasks}. For each task, operators were instructed to complete the objective while ensuring diverse and successful interaction patterns. We varied object positions and orientations across demonstrations to enhance data diversity.

% \textbf{Dataset Composition.}
The resulting dataset, which we term \textit{ForceVLA-Data}, comprises a total of 244 trajectories, amounting to 140 thousand synchronized timesteps. All sensor streams were synchronized based on timestamps. Images were resized to 480x640 pixels and normalized. Actions were represented as target TCP pose and gripper width. The \textit{ForceVLA-Data} dataset, along with our data collection code and processing scripts, will be made publicly available at \href{https://sites.google.com/view/forcevla2025/}{website} to facilitate further research in learning force-aware manipulation policies.

\sectionreducemargin{Experiments} \label{Sec: exp}

% This section details extensive real-world contact-rich manipulation experiments and analytical studies designed to empirically validate our proposed ForceVLA model. These investigations aim to address the following key research questions:

% \begin{itemize}[leftmargin=10pt]
% \item How effective is ForceVLA, and does it offer discernible improvements over approaches that directly incorporate force signals without our specialized fusion mechanism?
% \item To what extent can ForceVLA generalize its learned skills to variations in objects, environmental conditions, or task parameters not seen during training?
% \item Is the FVLMoE architecture effective in fusing modalities, and does its design approach optimality for enhancing performance in contact-rich tasks?
% \item Does the Mixture-of-Experts component effectively process disparate input modalities and exhibit appropriate, learned routing of information to specialized expert networks?
% \end{itemize}

This section presents a comprehensive suite of real-world contact-rich manipulation experiments and analytical studies to empirically validate the ForceVLA model. The evaluation is structured around four core research questions: (1) the overall effectiveness of ForceVLA compared to baselines that incorporate force without our specialized fusion mechanism; (2) the model’s ability to generalize across unseen object instances, environmental variations, and task conditions; (3) the efficacy of the proposed FVLMoE architecture in achieving optimal multimodal fusion for contact-rich manipulation; and (4) the ability of the Mixture-of-Experts module to appropriately process heterogeneous input modalities and learn meaningful routing behaviors across expert networks.

\subsectionreducemargin{Experimental Setups} \label{sec:evaluation_tasks}

\begin{figure}[t]
    \centering
    \includegraphics[width=0.95\columnwidth]{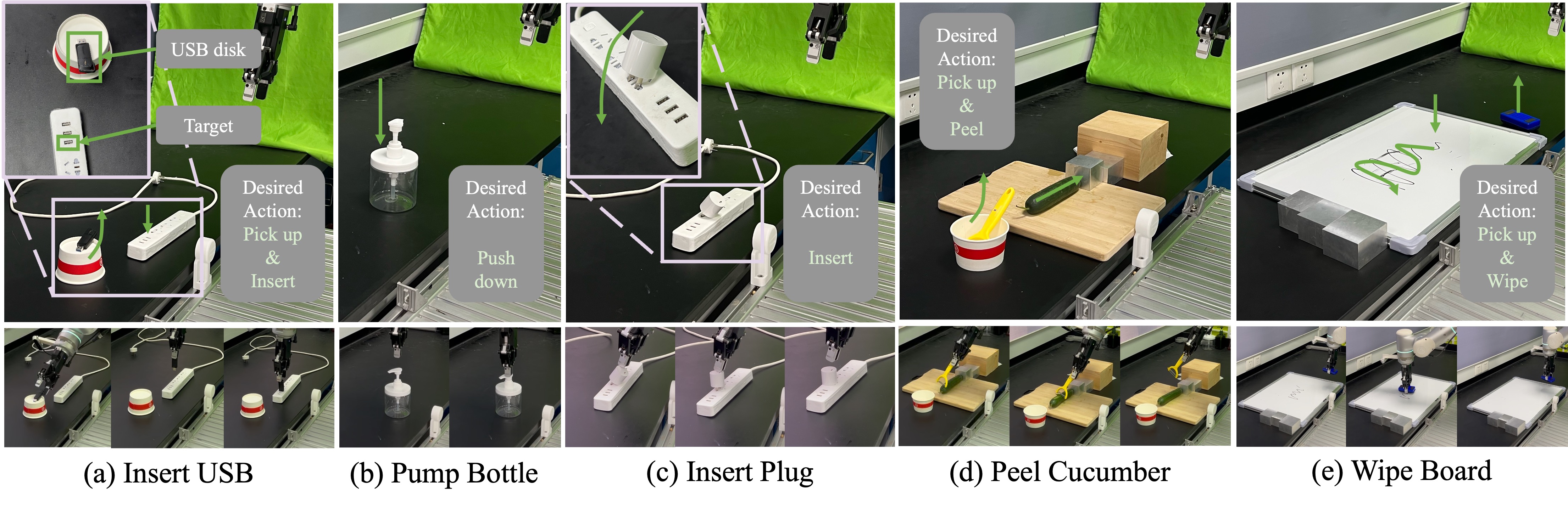}
    \caption{Overview of task setups used in evaluation. (a) Insert USB, (b) pump bottle, (c) insert plug, (d) peel cucumber, and (e) wipe board. These tasks span diverse contact dynamics and manipulation skills, from precise insertions to tool-mediated surface interactions.}
    \label{Fig: task_setup}
\vspace{-4mm}
\end{figure}

To evaluate the effectiveness of ForceVLA, we conducted experiments on five diverse contact-rich manipulation tasks: Bottle Pumping, Plug Insertion, USB Drive Insertion, Whiteboard Wiping, and Cucumber Peeling, as in Figure~\ref{Fig: task_setup}. These tasks were chosen to assess fine-grained control, adaptability to varied initial conditions, and the utility of multimodal feedback, particularly force sensing. Each task introduces unique physical challenges: Bottle Pumping requires precise vertical pressing; Plug and USB Drive Insertions involve accurate alignment and force-controlled insertion; Whiteboard Wiping demands smooth trajectory control and surface contact; and Cucumber Peeling tests the ability to apply and maintain controlled force during continuous surface interaction.

We trained ForceVLA using approximately 50 expert demonstrations per task. Evaluation was conducted over 20 trials each for the insertion and pumping tasks, 10 trials for the more time-consuming whiteboard task, and 15 trials for the cucumber peeling task, each involving 15 peeling strokes. Success was defined by task-specific criteria such as complete insertion, effective wiping motion, or substantial cumulative peel coverage. These tasks were designed to rigorously probe ForceVLA's capacity to model and control complex, uncertain dynamics through the integration of vision, language, and force modalities.

\begin{figure}[t]
    \centering
    \includegraphics[width=0.9\columnwidth]{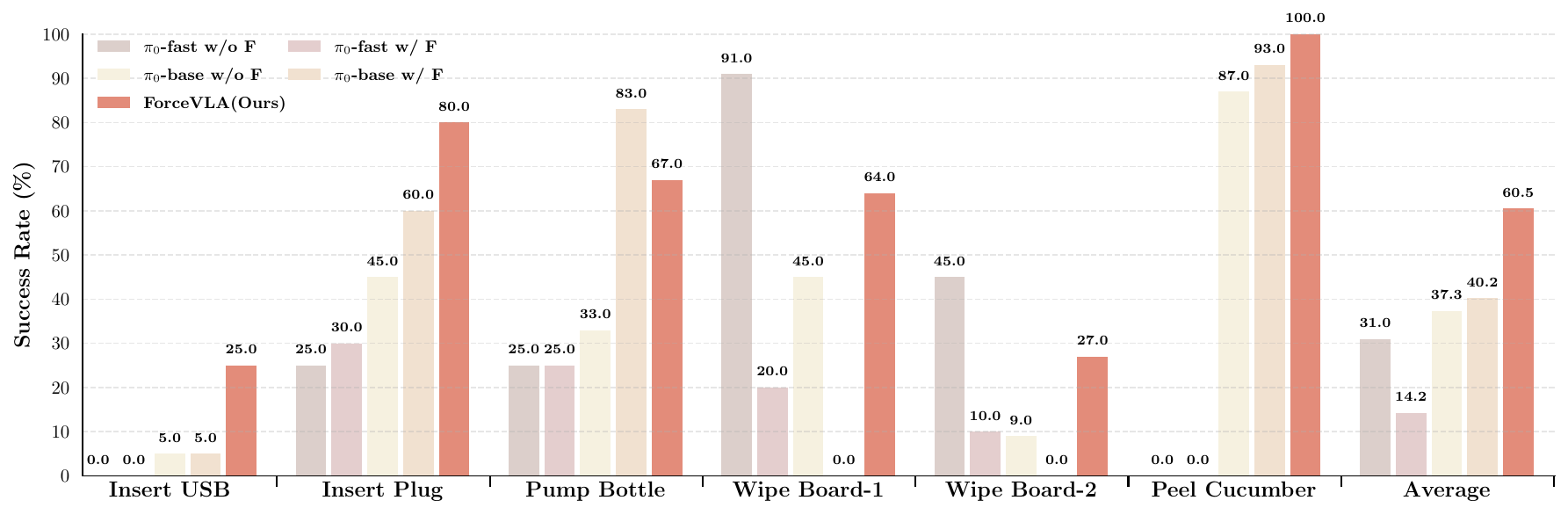}
    \caption{Main task success rates across different methods. ForceVLA significantly outperforms all baselines on five contact-rich tasks. Incorporating external force feedback improves performance for $\pi_0$-base model, while our method achieves the highest average success rate, demonstrating robust performance under complex interaction dynamics. ``Wipe Board-1'' indicates the success rate of successfully performing the wiping motion, while ``Wipe Board-2'' refers to the success rate of completely erasing the markings.}
    \label{Fig: main results table}
\vspace{-4mm}
\end{figure}

\textbf{Evaluation Metrics and Baselines.}
Model performance is primarily evaluated using the task success rate across all five challenging contact-rich manipulation tasks. For specific task like cucumber peeling, average peel length and minimum peeling times are also reported to provide a more nuanced assessment. To contextualize the performance of our proposed ForceVLA model, we compare it against several carefully selected baselines derived from the state-of-the-art $\pi_0$\cite{pi_0} architecture, which serves as our foundational model. The specific variants include $\pi_0$-base\cite{pi_0} w/o F (standard $\pi_0$ without force input), $\pi_0$-base\cite{pi_0} w/ F ($\pi_0$ with force signals directly concatenated to state inputs), and corresponding $\pi_0$-fast\cite{fastpi} configurations ( w/o F and w/ F), representing potentially faster alternatives. The selection of $\pi_0$-base\cite{pi_0} allows comparison with a strong existing VLA method, while the `inputForce` variants are crucial for demonstrating the efficacy of our FVLMoE fusion strategy over simpler force integration approaches.

% Compared Methods: $\pi_0$-base w/o F, $\pi_0$-base w/ F, $\pi_0$-fast w/o F, $\pi_0$-fast w/ F.
% Why were these chosen? $\pi_0$-base is a SOTA method and our baseline. The 'inputForce' variants are used to demonstrate the necessity of our MoE-based fusion over simpler force integration.
% %
% placeholder. placeholder. placeholder. placeholder. placeholder. placeholder. placeholder. placeholder. placeholder. placeholder. placeholder. placeholder.

% \textbf{Implementation Details.}
% Important hyperparameter settings, training strategies, hardware environment, etc.

\subsectionreducemargin{Main Results}

\begin{wraptable}{R}{0.48\textwidth}
\vspace{-5mm}
  \centering
  \caption{Performance of cucumber peeling.}
  % Higher average peel length and lower minimum strokes indicate better performance.
  % Best values are in \textbf{bold}.
  \label{tab:cucumber_peeling_wrapped_arrows} % Updated label
  \renewcommand{\theadfont}{\small\bfseries} % Make headers small and bold
  \renewcommand{\theadalign}{cc} % Center-align content within \thead cells
  \setlength{\tabcolsep}{3pt} % Reduce space between columns (default is 6pt)
  \small % Make the font size of the table content smaller
  \begin{tabular}{lcc} 
  \toprule
  \thead{Model} & \thead{Avg. Peel\\Length (cm) $\uparrow$} & \thead{Min. Strokes\\to Clean $\downarrow$} \\ % Added arrows
  \midrule
  $\pi_0$-base\cite{pi_0} w/o F    & 10.27                 & 14                    \\
  $\pi_0$-base\cite{pi_0} w/ F   & 13.17                 & 10                    \\
  \textbf{ForceVLA (Ours)} & \textbf{14.12}        & \textbf{7}            \\
  \bottomrule
  \end{tabular}
\vspace{-3mm}
\end{wraptable}

% This section presents the primary findings from our extensive experiments, evaluating ForceVLA's performance on five challenging contact-rich manipulation tasks against various baseline models.

\textbf{Overall Performance.}
As demonstrated in Figure~\ref{Fig: main results table}, ForceVLA achieves an average success rate of 60.5\% across all five tasks, significantly outperforming all baseline configurations. Compared to the standard $\pi_0$-base model without force feedback ($\pi_0$-base w/ F), which achieved an average of 37.3\%, ForceVLA shows an improvement of 23.2\%. This highlights the substantial benefit of incorporating and effectively processing multimodal information, including force. Table~\ref{tab:cucumber_peeling_wrapped_arrows} further highlights ForceVLA's superior performance on the intricate \textit{cucumber peeling} task. Our model excelled on both key metrics: it achieved the longest average peel length per stroke (14.12\,cm~$\uparrow$), indicating better ability to execute high-fidelity surface manipulation through stable tool orientation, adaptive contouring, and sustained surface contact compared to both $\pi_0$-base w/ F (13.17\,cm) and $\pi_0$-base w/o F (10.27\,cm). Concurrently, ForceVLA demonstrated superior overall efficiency by requiring the minimum number of strokes (7~$\downarrow$) to achieve a substantially peeled cucumber, significantly fewer than the 10 and 14 strokes needed by $\pi_0$-base w/ F and $\pi_0$-base w/o F, respectively. These combined results underscore ForceVLA's proficiency in maintaining consistent, effective tool-surface interaction and executing efficient, goal-directed motions in tasks demanding continuous and precise force modulation.

% \textbf{Overall Performance of ForceVLA.}
% ForceVLA achieves an average success rate of 60.5\% across five tasks (Figure~\ref{Fig: main results table}), outperforming the best baseline ($\pi_0$-base w/ F: 37.3\%) by 23.2\%. On the challenging \textit{cucumber peeling} task (Table~\ref{tab:cucumber_peeling_wrapped_arrows}), ForceVLA delivered the longest average peel length (14.12\,cm) and required the fewest strokes (7) to reach a substantially peeled outcome, outperforming both $\pi_0$-base w/ F (13.17\,cm, 10 strokes) and w/o F (10.27\,cm, 14 strokes). These results highlight ForceVLA’s capacity for precise, force-aware surface manipulation and efficient task execution.

% \textbf{Impact of Force Modality and FVLMoE Fusion.}
% The importance of force perception is evident when comparing $\pi_0$-base w/o Force with $\pi_0$-base w/ Force. $\pi_0$-base w/ Force model simply concatenates force signals with state inputs, consequently improves the average success rate to 40.2\%, affirming the value of force data. However, ForceVLA further advances this performance to 60.5\%. This additional gain of 20.3\% underscores the critical contribution of the FVLMoE module in effectively integrating force with visual and linguistic context, surpassing simpler fusion approaches.

\textbf{Effectiveness of Force Fusion via FVLMoE.}
Introducing raw force signals into $\pi_0$-base boosts performance from 37.3\% to 40.2\%, confirming the utility of force feedback. However, ForceVLA surpasses both with 60.5\%, indicating that effective fusion—enabled by our FVLMoE module—is essential for fully leveraging tactile information. This demonstrates that beyond the presence of force data, how it is integrated is critical to performance gains.

\textbf{Selection of $\pi_0$-base and $\pi_0$-fast model.}
For our foundational baseline, we evaluated $\pi_0$-base and $\pi_0$-fast variants.
The $\pi_0$-base architecture demonstrated superior overall performance: $\pi_0$-base w/ F (40.2\%) and $\pi_0$-base w/o F (37.3\%) significantly outperformed $\pi_0$-fast w/ F (14.2\%) and $\pi_0$-fast w/o F (31.0\%).
While $\pi_0$-fast variants exhibited a comparative advantage solely on the whiteboard wiping task—potentially due to a simpler action generation mechanism being more attuned to such motions, the $\pi_0$-fast architecture's performance notably degraded (from 31.0\% to 14.2\%) when raw force input was directly added.
We attribute this sensitivity to its highly optimized and compact token space, which is likely disrupted by naively projected force tokens lacking corresponding large-scale pre-training.
Conversely, $\pi_0$-base modestly benefited from direct force input, with its larger representational capacity presumably allowing for partial utilization of these new sensory signals.
Given its superior aggregate performance and more robust handling of naive force integration, $\pi_0$-base was selected as the primary baseline for developing and evaluating ForceVLA.

% \textbf{Baseline Selection: $\pi_0$-base vs. $\pi_0$-fast.}
% Among baselines, $\pi_0$-base showed more stable and higher performance than $\pi_0$-fast: 40.2\% vs. 14.2\% (with force) and 37.3\% vs. 31.0\% (w/o force). Although $\pi_0$-fast performed better on the whiteboard task, its compact token space proved highly sensitive to force input. In contrast, $\pi_0$-base better tolerated naive force fusion, justifying its selection as the foundation for developing ForceVLA.

\subsection{Model Generalization}

\begin{figure}[t]
    \centering
    \includegraphics[width=0.9\columnwidth]{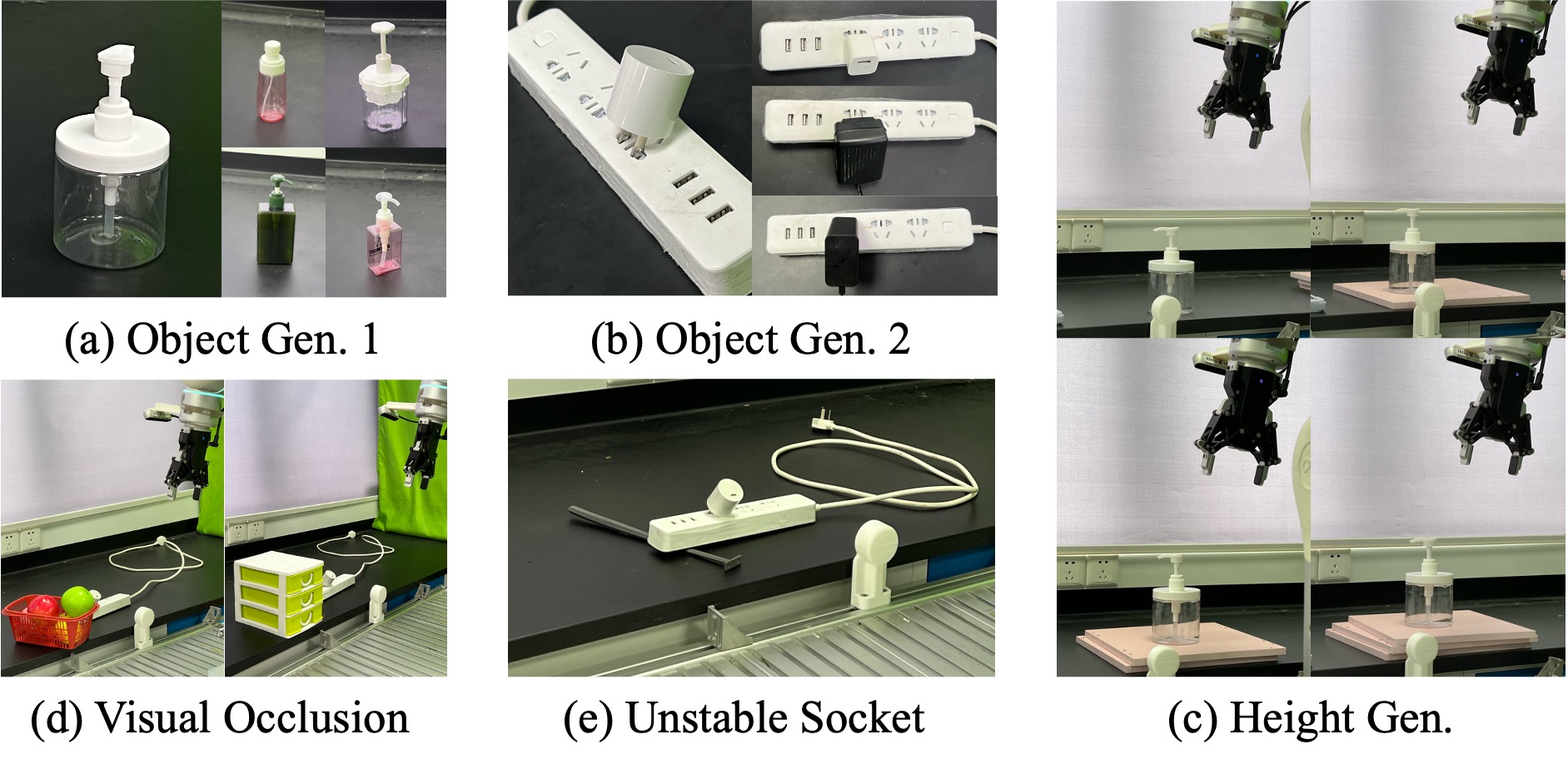}
    \caption{Variants of generalization settings used in our experiments. (a–b) Different object geometries; (c) variation in socket height; (d) partial visual occlusion; (e) unstable socket conditions. These scenarios evaluate robustness under diverse physical and perceptual perturbations.}
    \label{Fig: generalization}
\vspace{-2mm}
\end{figure}

To evaluate ForceVLA’s generalization capabilities, we designed five experimental settings with increasing task variability and physical uncertainty, as illustrated in Figure~\ref{Fig: generalization}. These settings include: (1) \textbf{Object Gen. 1}, which varies the bottle type in the bottle pumping task; (2) \textbf{Object Gen. 2}, which changes the plug type in the plug insertion task; (3) \textbf{Height Gen.}, which adjusts the initial bottle height and measures success under torque limits; (4) \textbf{Visual Occlusion}, where parts of the plug and socket are obscured; and (5) \textbf{Unstable Socket}, introducing physical instability via clutter beneath the socket. These variations test both perceptual robustness and physical adaptability, with results summarized in Table~\ref{tab:generalization_results}.

Across all settings, ForceVLA exhibited superior generalization, particularly in scenarios requiring fine physical interaction. In Object Gen. 1, it achieved an 80.00\% success rate, outperforming baselines that lacked force input or processed it naively. In the Height Gen. setting, ForceVLA effectively scaled its interaction forces to variable depths, avoiding torque limit violations seen in other models. Furthermore, ForceVLA maintained high success under visual degradation (90.00\% in Visual Occlusion), reflecting its reliance on multimodal feedback beyond visual cues. These results underscore the critical role of the proposed FVLMoE architecture in intelligently integrating force information-not just for sensing contact, but for modulating action in response to dynamic physical conditions—enabling more versatile and robust robotic manipulation.

\begin{table}[t]
  \caption{Success rates (\%) of different models under various experimental conditions. Maximum values in each column are highlighted in \textbf{bold}; second-best values are \underline{underlined}.}
  \label{tab:generalization_results} % Updated label
  \centering
  \renewcommand{\theadalign}{cc} % Center align content within \thead cells
  \setlength{\tabcolsep}{3pt} % Further reduced column separation slightly for the new column
  \begin{tabular}{lcccccc} % Changed from lccccc to lcccccc for the new column
    \toprule
    \textbf{Model} & \textbf{\makecell{Object\\Gen. 1}} & \textbf{\makecell{Object\\Gen. 2}} & \textbf{\makecell{Height\\Gen.}} & \textbf{\makecell{Visual\\Occlusion}} & \textbf{\makecell{Unstable\\Socket}} & \textbf{\makecell{Average}} \\ % Added new header
    \midrule
    $\pi_0$-base\cite{pi_0} w/o F & \underline{48.00\%} & 10.00\%          & 66.67\%          & \underline{60.00\%} & 10.00\%          & 38.93\% \\
    $\pi_0$-base\cite{pi_0} w/ F  & 32.00\%             & 10.00\%          & \underline{77.78\%} & 30.00\%             & 10.00\%          & 31.96\% \\
    $\pi_0$-fast\cite{fastpi} w/o F & \textbf{80.00\%}    & \underline{35.00\%} & \textbf{88.89\%}    & 50.00\%             & 10.00\%          & \underline{52.78\%} \\
    $\pi_0$-fast\cite{fastpi} w/ F  & 32.00\%             & 5.00\%           & 44.44\%             & 50.00\%             & \textbf{30.00\%} & 32.29\% \\
    ForceVLA (Ours)    & \textbf{80.00\%}    & \textbf{40.00\%} & \textbf{88.89\%}    & \textbf{90.00\%}    & \underline{20.00\%} & \textbf{63.78\%} \\
    \bottomrule
  \end{tabular}
\vspace{-4mm}
\end{table}

\subsectionreducemargin{Ablation Studies} \label{Sec: ablation}

\begin{wraptable}{R}{0.4\textwidth}
\vspace{-4mm}
  \centering
  \caption{Ablation Results}
  \label{Tab: ablation}
  \begin{tabular}{lc}
  \toprule
  Model & Success Rate \\
  \midrule
  baseline\cite{pi_0} & 45\% \\
  linear before VLM & 55\% \\
  MoE before VLM & 0 \\
  concate after VLM & 60\% \\
  \textbf{ForceVLA (Ours)} & \textbf{80}\% \\
  \bottomrule
  \end{tabular}
\vspace{-4mm}
\end{wraptable}

To validate the architectural design of \textit{ForceVLA}, particularly the integration of force feedback, we conducted comprehensive ablation studies shown in Table~\ref{Tab: ablation}. We compared early, late, and our proposed fusion strategy. Early fusion methods, such as ``linear before VLM'' and ``MoE before VLM,'' which inject force data prior to the visual-language model (VLM), significantly degraded performance. Notably, the MoE-based early fusion failed entirely (0\% success rate), highlighting that altering the input representations of a pretrained VLM disrupts its learned feature distributions and undermines its capacity to process visual-linguistic signals effectively.

Late fusion strategies fared better. The ``concatenate after VLM'' method, equivalent to a basic baseline where force features are appended at the decoding stage, improved success to 60\%—demonstrating the utility of force sensing. However, our proposed \textbf{ForceVLA} architecture achieved a markedly higher 80\% success rate. By introducing force features after the VLM’s core encoding and using the FVLMoE module for adaptive fusion, ForceVLA enables specialized routing and deeper multimodal interaction. These results confirm two core design insights: force should be introduced post-VLM to preserve pretrained representations, and sophisticated fusion (via FVLMoE) is essential to fully leverage force in guiding contact-rich robotic behavior.

\subsectionreducemargin{Visualization and Case Studies}

\begin{figure}[t]
    \centering
    \includegraphics[width=0.97\columnwidth]{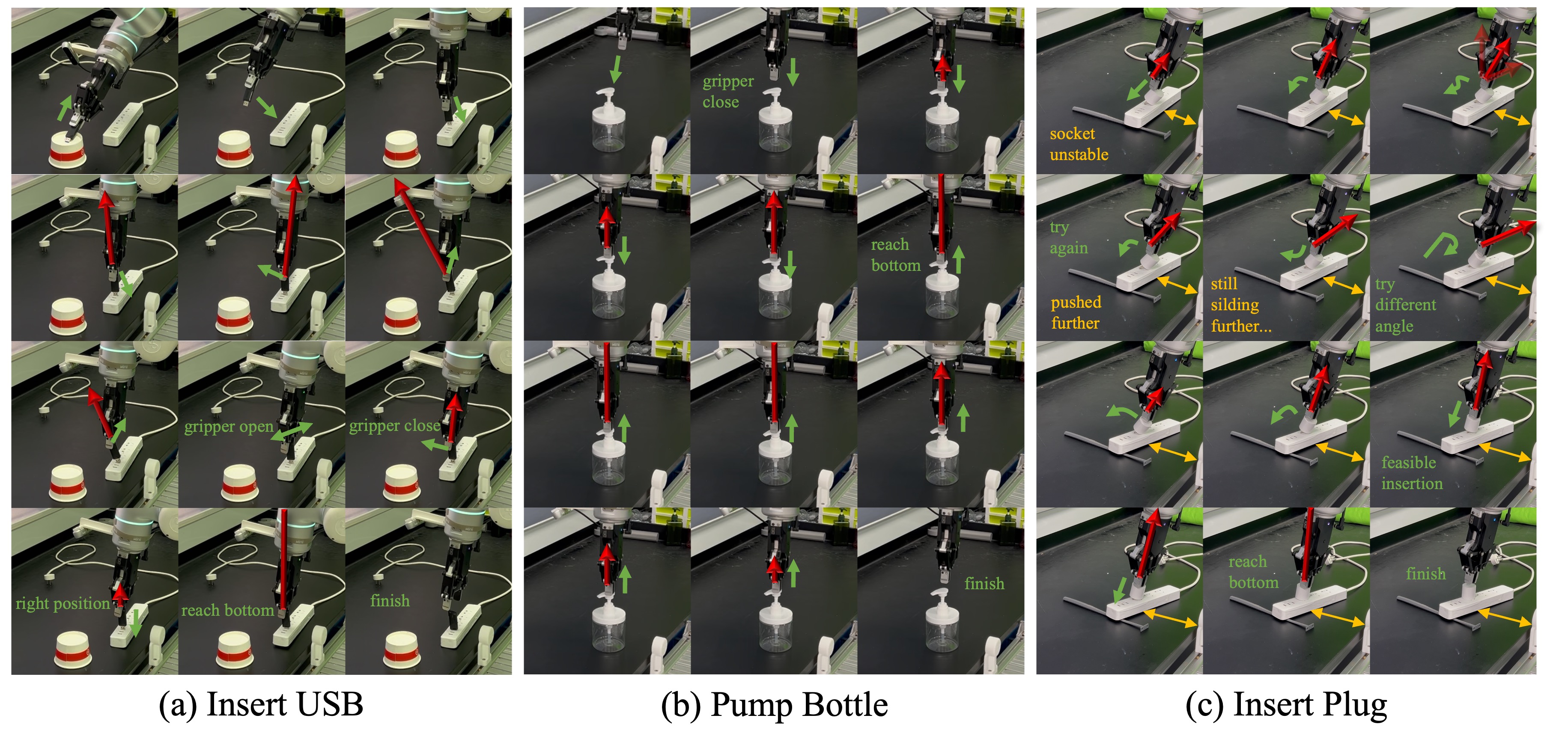}
    \caption{Trajectory visualizations across tasks and conditions. (a) USB insertion, (b) bottle pumping, and (c) plug insertion under stable and unstable socket conditions. Each sequence illustrates how ForceVLA adapts its actions in response to contact dynamics, retrying or adjusting pose when failures occur, ultimately achieving successful task completion.}
    \label{Fig: visualization}
\vspace{-4mm}
\end{figure}

Figure~\ref{Fig: visualization} illustrates ForceVLA’s ability to adapt motion in response to contact feedback during complex manipulation tasks. In the USB insertion task, when initial attempts failed due to misalignment, ForceVLA re-oriented or re-grasped the drive to achieve successful insertion—behaviors absent in baseline models, which repeated failed motions or applied excessive force. Similarly, in the ``Unstable Socket'' scenario (Figure~\ref{Fig: visualization}c), ForceVLA maintained compliant control as the socket shifted, dynamically adjusting the plug’s pose to complete insertion, while baselines lost tracking and failed. These examples highlight a key insight: simply adding force input does not ensure closed-loop adaptation. ForceVLA’s FVLMoE module enables deep fusion of force, vision, and language, supporting precise, context-aware control and robust generalization under dynamic physical conditions.

% \begin{figure}[htbp]
%     \centering
%     \includegraphics[width=0.90\columnwidth]{figures/expert_trend_insert_plug.pdf}
%     \caption{Top-1 routing probabilities over time for insert plug and pump bottle tasks}
%     \label{Fig: router_visualization}
% \vspace{-4mm}
% \end{figure}
\sectionreducemargin{Conclusion} \label{Sec: conclusion}

In conclusion, we propose ForceVLA, a framework that bridges the gap between high-level modality (vision/language) and low-level physical sensing (force) for contact-rich manipulation. At its core, ForceVLA introduces \textbf{FVLMoE}, a Mixture-of-Experts module that dynamically fuses visual, linguistic, and force modalities to enable fine-grained, context-aware control. Our experiments across five challenging tasks show that ForceVLA significantly outperforms strong $\pi_0$-based baselines, achieving an average success rate improvement of 23.2\% and up to 80\% success on individual tasks. Ablation studies further validate the benefits of late-stage force fusion and expert routing. We also contribute \textbf{ForceVLA-Data}, a new dataset for multimodal contact-rich manipulation. Through the co-design of our architectural approach and dataset, we demonstrate meaningful progress toward VLA systems that exhibit greater adaptive behavior and physical intelligence in manipulation scenarios.

\textbf{Limitation.} 
Firstly, ForceVLA currently utilizes estimated external wrench values. While this approach has proven effective, these estimations may not always capture the full precision afforded by direct high-fidelity measurements, particularly in scenarios demanding extreme haptic sensitivity. Potential enhancements include exploring the integration of superior sensors or advanced calibration techniques to further refine fine-grained control capabilities. Secondly, ForceVLA's experimental validation was predominantly conducted on robotic platforms with integrated, and typically high-cost, force-torque sensing, which can naturally limit broader accessibility. To promote wider practical deployment and help democratize force-aware manipulation research, we are actively assessing ForceVLA's adaptability and performance on more common, lower-cost platforms equipped with external or retrofitted force sensors.

% \clearpage

{
\small

\bibliographystyle{unsrt}
\bibliography{reference}

\begin{thebibliography}{10}

\bibitem{droid}
Alexander Khazatsky, Karl Pertsch, Suraj Nair, Ashwin Balakrishna, Sudeep Dasari, Siddharth Karamcheti, Soroush Nasiriany, Mohan~Kumar Srirama, Lawrence~Yunliang Chen, Kirsty Ellis, et~al.
\newblock Droid: A large-scale in-the-wild robot manipulation dataset.
\newblock {\em arXiv preprint arXiv:2403.12945}, 2024.

\bibitem{openx-embodiment}
Abby O’Neill, Abdul Rehman, Abhiram Maddukuri, Abhishek Gupta, Abhishek Padalkar, Abraham Lee, Acorn Pooley, Agrim Gupta, Ajay Mandlekar, Ajinkya Jain, et~al.
\newblock Open x-embodiment: Robotic learning datasets and rt-x models: Open x-embodiment collaboration 0.
\newblock In {\em 2024 IEEE International Conference on Robotics and Automation (ICRA)}, pages 6892--6903. IEEE, 2024.

\bibitem{Bridgedatav2}
Homer~Rich Walke, Kevin Black, Tony~Z Zhao, Quan Vuong, Chongyi Zheng, Philippe Hansen-Estruch, Andre~Wang He, Vivek Myers, Moo~Jin Kim, Max Du, et~al.
\newblock Bridgedata v2: A dataset for robot learning at scale.
\newblock In {\em Conference on Robot Learning}, pages 1723--1736. PMLR, 2023.

\bibitem{rt1}
Anthony Brohan, Noah Brown, Justice Carbajal, Yevgen Chebotar, Joseph Dabis, Chelsea Finn, Keerthana Gopalakrishnan, Karol Hausman, Alex Herzog, Jasmine Hsu, et~al.
\newblock Rt-1: Robotics transformer for real-world control at scale.
\newblock {\em arXiv preprint arXiv:2212.06817}, 2022.

\bibitem{team2024octo}
Octo~Model Team, Dibya Ghosh, Homer Walke, Karl Pertsch, Kevin Black, Oier Mees, Sudeep Dasari, Joey Hejna, Tobias Kreiman, Charles Xu, et~al.
\newblock Octo: An open-source generalist robot policy.
\newblock {\em arXiv preprint arXiv:2405.12213}, 2024.

\bibitem{brohan2023rt}
Anthony Brohan, Noah Brown, Justice Carbajal, Yevgen Chebotar, Xi~Chen, Krzysztof Choromanski, Tianli Ding, Danny Driess, Avinava Dubey, Chelsea Finn, et~al.
\newblock Rt-2: Vision-language-action models transfer web knowledge to robotic control.
\newblock {\em arXiv preprint arXiv:2307.15818}, 2023.

\bibitem{radford2021learning}
Alec Radford, Jong~Wook Kim, Chris Hallacy, Aditya Ramesh, Gabriel Goh, Sandhini Agarwal, Girish Sastry, Amanda Askell, Pamela Mishkin, Jack Clark, et~al.
\newblock Learning transferable visual models from natural language supervision.
\newblock In {\em International conference on machine learning}, pages 8748--8763. PmLR, 2021.

\bibitem{zhuang2025math}
Wenwen Zhuang, Xin Huang, Xiantao Zhang, and Jin Zeng.
\newblock Math-puma: Progressive upward multimodal alignment to enhance mathematical reasoning.
\newblock In {\em Proceedings of the AAAI Conference on Artificial Intelligence}, volume~39, pages 26183--26191, 2025.

\bibitem{openvla}
Moo~Jin Kim, Karl Pertsch, Siddharth Karamcheti, Ted Xiao, Ashwin Balakrishna, Suraj Nair, Rafael Rafailov, Ethan Foster, Grace Lam, Pannag Sanketi, et~al.
\newblock Openvla: An open-source vision-language-action model.
\newblock {\em arXiv preprint arXiv:2406.09246}, 2024.

\bibitem{pi_0}
Kevin Black, Noah Brown, Danny Driess, Adnan Esmail, Michael Equi, Chelsea Finn, Niccolo Fusai, Lachy Groom, Karol Hausman, Brian Ichter, et~al.
\newblock $\pi_0$: A vision-language-action flow model for general robot control.
\newblock {\em arXiv preprint arXiv:2410.24164}, 2024.

\bibitem{paligemma}
Lucas Beyer, Andreas Steiner, Andr{\'e}~Susano Pinto, Alexander Kolesnikov, Xiao Wang, Daniel Salz, Maxim Neumann, Ibrahim Alabdulmohsin, Michael Tschannen, Emanuele Bugliarello, et~al.
\newblock Paligemma: A versatile 3b vlm for transfer.
\newblock {\em arXiv preprint arXiv:2407.07726}, 2024.

\bibitem{contact1}
Xiang Zhang, Changhao Wang, Lingfeng Sun, Zheng Wu, Xinghao Zhu, and Masayoshi Tomizuka.
\newblock Efficient sim-to-real transfer of contact-rich manipulation skills with online admittance residual learning.
\newblock In {\em Conference on Robot Learning}, pages 1621--1639. PMLR, 2023.

\bibitem{forge}
Michael Noseworthy, Bingjie Tang, Bowen Wen, Ankur Handa, Chad Kessens, Nicholas Roy, Dieter Fox, Fabio Ramos, Yashraj Narang, and Iretiayo Akinola.
\newblock Forge: Force-guided exploration for robust contact-rich manipulation under uncertainty.
\newblock {\em IEEE Robotics and Automation Letters}, 2025.

\bibitem{kim2015multimodal}
Sung~Soo Kim, Manuel Gomez-Ramirez, Pramodsingh~H Thakur, and Steven~S Hsiao.
\newblock Multimodal interactions between proprioceptive and cutaneous signals in primary somatosensory cortex.
\newblock {\em Neuron}, 86(2):555--566, 2015.

\bibitem{openvla-oft}
Moo~Jin Kim, Chelsea Finn, and Percy Liang.
\newblock Fine-tuning vision-language-action models: Optimizing speed and success.
\newblock {\em arXiv preprint arXiv:2502.19645}, 2025.

\bibitem{vlas}
Wei Zhao, Pengxiang Ding, Min Zhang, Zhefei Gong, Shuanghao Bai, Han Zhao, and Donglin Wang.
\newblock Vlas: Vision-language-action model with speech instructions for customized robot manipulation.
\newblock {\em arXiv preprint arXiv:2502.13508}, 2025.

\bibitem{robomamba}
Jiaming Liu, Mengzhen Liu, Zhenyu Wang, Lily Lee, Kaichen Zhou, Pengju An, Senqiao Yang, Renrui Zhang, Yandong Guo, and Shanghang Zhang.
\newblock Robomamba: Multimodal state space model for efficient robot reasoning and manipulation.
\newblock {\em arXiv preprint arXiv:2406.04339}, 2024.

\bibitem{vima}
Yunfan Jiang, Agrim Gupta, Zichen Zhang, Guanzhi Wang, Yongqiang Dou, Yanjun Chen, Li~Fei-Fei, Anima Anandkumar, Yuke Zhu, and Linxi Fan.
\newblock Vima: General robot manipulation with multimodal prompts.
\newblock {\em arXiv preprint arXiv:2210.03094}, 2(3):6, 2022.

\bibitem{gr1}
Johan Bjorck, Fernando Casta{\~n}eda, Nikita Cherniadev, Xingye Da, Runyu Ding, Linxi Fan, Yu~Fang, Dieter Fox, Fengyuan Hu, Spencer Huang, et~al.
\newblock Gr00t n1: An open foundation model for generalist humanoid robots.
\newblock {\em arXiv preprint arXiv:2503.14734}, 2025.

\bibitem{interleave}
Cunxin Fan, Xiaosong Jia, Yihang Sun, Yixiao Wang, Jianglan Wei, Ziyang Gong, Xiangyu Zhao, Masayoshi Tomizuka, Xue Yang, Junchi Yan, et~al.
\newblock Interleave-vla: Enhancing robot manipulation with interleaved image-text instructions.
\newblock {\em arXiv preprint arXiv:2505.02152}, 2025.

\bibitem{pi0.5}
Physical Intelligence, Kevin Black, Noah Brown, James Darpinian, Karan Dhabalia, Danny Driess, Adnan Esmail, Michael Equi, Chelsea Finn, Niccolo Fusai, et~al.
\newblock $\pi_0.5$: a vision-language-action model with open-world generalization.
\newblock {\em arXiv preprint arXiv:2504.16054}, 2025.

\bibitem{cot-vla}
Qingqing Zhao, Yao Lu, Moo~Jin Kim, Zipeng Fu, Zhuoyang Zhang, Yecheng Wu, Zhaoshuo Li, Qianli Ma, Song Han, Chelsea Finn, et~al.
\newblock Cot-vla: Visual chain-of-thought reasoning for vision-language-action models.
\newblock {\em arXiv preprint arXiv:2503.22020}, 2025.

\bibitem{dexvla}
Junjie Wen, Yichen Zhu, Jinming Li, Zhibin Tang, Chaomin Shen, and Feifei Feng.
\newblock Dexvla: Vision-language model with plug-in diffusion expert for general robot control.
\newblock {\em arXiv preprint arXiv:2502.05855}, 2025.

\bibitem{robodual}
Qingwen Bu, Hongyang Li, Li~Chen, Jisong Cai, Jia Zeng, Heming Cui, Maoqing Yao, and Yu~Qiao.
\newblock Towards synergistic, generalized, and efficient dual-system for robotic manipulation.
\newblock {\em arXiv preprint arXiv:2410.08001}, 2024.

\bibitem{fastpi}
Karl Pertsch, Kyle Stachowicz, Brian Ichter, Danny Driess, Suraj Nair, Quan Vuong, Oier Mees, Chelsea Finn, and Sergey Levine.
\newblock Fast: Efficient action tokenization for vision-language-action models.
\newblock {\em arXiv preprint arXiv:2501.09747}, 2025.

\bibitem{pointvla}
Chengmeng Li, Junjie Wen, Yan Peng, Yaxin Peng, Feifei Feng, and Yichen Zhu.
\newblock Pointvla: Injecting the 3d world into vision-language-action models.
\newblock {\em arXiv preprint arXiv:2503.07511}, 2025.

\bibitem{dp}
Cheng Chi, Zhenjia Xu, Siyuan Feng, Eric Cousineau, Yilun Du, Benjamin Burchfiel, Russ Tedrake, and Shuran Song.
\newblock Diffusion policy: Visuomotor policy learning via action diffusion.
\newblock {\em The International Journal of Robotics Research}, page 02783649241273668, 2023.

\bibitem{prediction}
Yanjiang Guo, Yucheng Hu, Jianke Zhang, Yen-Jen Wang, Xiaoyu Chen, Chaochao Lu, and Jianyu Chen.
\newblock Prediction with action: Visual policy learning via joint denoising process.
\newblock In {\em The Thirty-eighth Annual Conference on Neural Information Processing Systems}, 2024.

\bibitem{chatvla}
Zhongyi Zhou, Yichen Zhu, Minjie Zhu, Junjie Wen, Ning Liu, Zhiyuan Xu, Weibin Meng, Ran Cheng, Yaxin Peng, Chaomin Shen, et~al.
\newblock Chatvla: Unified multimodal understanding and robot control with vision-language-action model.
\newblock {\em arXiv preprint arXiv:2502.14420}, 2025.

\bibitem{hybridvla}
Jiaming Liu, Hao Chen, Pengju An, Zhuoyang Liu, Renrui Zhang, Chenyang Gu, Xiaoqi Li, Ziyu Guo, Sixiang Chen, Mengzhen Liu, et~al.
\newblock Hybridvla: Collaborative diffusion and autoregression in a unified vision-language-action model.
\newblock {\em arXiv preprint arXiv:2503.10631}, 2025.

\bibitem{tacdiffusion}
Yansong Wu, Zongxie Chen, Fan Wu, Lingyun Chen, Liding Zhang, Zhenshan Bing, Abdalla Swikir, Sami Haddadin, and Alois Knoll.
\newblock Tacdiffusion: Force-domain diffusion policy for precise tactile manipulation.
\newblock {\em arXiv preprint arXiv:2409.11047}, 2024.

\bibitem{adaptiveCP}
Yifan Hou, Zeyi Liu, Cheng Chi, Eric Cousineau, Naveen Kuppuswamy, Siyuan Feng, Benjamin Burchfiel, and Shuran Song.
\newblock Adaptive compliance policy: Learning approximate compliance for diffusion guided control.
\newblock {\em arXiv preprint arXiv:2410.09309}, 2024.

\bibitem{forcemimic}
Wenhai Liu, Junbo Wang, Yiming Wang, Weiming Wang, and Cewu Lu.
\newblock Forcemimic: Force-centric imitation learning with force-motion capture system for contact-rich manipulation.
\newblock {\em arXiv preprint arXiv:2410.07554}, 2024.

\bibitem{force-aware}
Zihao He, Hongjie Fang, Jingjing Chen, Hao-Shu Fang, and Cewu~Lu Foar.
\newblock Force-aware reactive policy for contact-rich robotic manipulation.
\newblock {\em arXiv preprint arXiv:2411.15753}, 2024.

\bibitem{towards}
William Xie and Nikolaus Correll.
\newblock Towards forceful robotic foundation models: a literature survey.
\newblock {\em arXiv preprint arXiv:2504.11827}, 2025.

\bibitem{tla}
Peng Hao, Chaofan Zhang, Dingzhe Li, Xiaoge Cao, Xiaoshuai Hao, Shaowei Cui, and Shuo Wang.
\newblock Tla: Tactile-language-action model for contact-rich manipulation.
\newblock {\em arXiv preprint arXiv:2503.08548}, 2025.

\bibitem{tac-man}
Zihang Zhao, Yuyang Li, Wanlin Li, Zhenghao Qi, Lecheng Ruan, Yixin Zhu, and Kaspar Althoefer.
\newblock Tac-man: Tactile-informed prior-free manipulation of articulated objects.
\newblock {\em IEEE Transactions on Robotics}, 2024.

\bibitem{impact}
Yiyang Ling, Karan Owalekar, Oluwatobiloba Adesanya, Erdem B{\i}y{\i}k, and Daniel Seita.
\newblock Impact: Intelligent motion planning with acceptable contact trajectories via vision-language models.
\newblock {\em arXiv preprint arXiv:2503.10110}, 2025.

\bibitem{should}
Huaijiang Zhu, Tong Zhao, Xinpei Ni, Jiuguang Wang, Kuan Fang, Ludovic Righetti, and Tao Pang.
\newblock Should we learn contact-rich manipulation policies from sampling-based planners?
\newblock {\em IEEE Robotics and Automation Letters}, 2025.

\bibitem{Outrageously}
Noam Shazeer, Azalia Mirhoseini, Krzysztof Maziarz, Andy Davis, Quoc Le, Geoffrey Hinton, and Jeff Dean.
\newblock Outrageously large neural networks: The sparsely-gated mixture-of-experts layer.
\newblock {\em arXiv preprint arXiv:1701.06538}, 2017.

\bibitem{switch}
William Fedus, Barret Zoph, and Noam Shazeer.
\newblock Switch transformers: Scaling to trillion parameter models with simple and efficient sparsity.
\newblock {\em Journal of Machine Learning Research}, 23(120):1--39, 2022.

\bibitem{gshard}
Dmitry Lepikhin, HyoukJoong Lee, Yuanzhong Xu, Dehao Chen, Orhan Firat, Yanping Huang, Maxim Krikun, Noam Shazeer, and Zhifeng Chen.
\newblock Gshard: Scaling giant models with conditional computation and automatic sharding.
\newblock {\em arXiv preprint arXiv:2006.16668}, 2020.

\bibitem{glam}
Nan Du, Yanping Huang, Andrew~M Dai, Simon Tong, Dmitry Lepikhin, Yuanzhong Xu, Maxim Krikun, Yanqi Zhou, Adams~Wei Yu, Orhan Firat, et~al.
\newblock Glam: Efficient scaling of language models with mixture-of-experts.
\newblock In {\em International conference on machine learning}, pages 5547--5569. PMLR, 2022.

\bibitem{sparsemoe}
Carlos Riquelme, Joan Puigcerver, Basil Mustafa, Maxim Neumann, Rodolphe Jenatton, Andr{\'e} Susano~Pinto, Daniel Keysers, and Neil Houlsby.
\newblock Scaling vision with sparse mixture of experts.
\newblock {\em Advances in Neural Information Processing Systems}, 34:8583--8595, 2021.

\bibitem{st-moe}
Barret Zoph, Irwan Bello, Sameer Kumar, Nan Du, Yanping Huang, Jeff Dean, Noam Shazeer, and William Fedus.
\newblock St-moe: Designing stable and transferable sparse expert models.
\newblock {\em arXiv preprint arXiv:2202.08906}, 2022.

\bibitem{limoe}
Basil Mustafa, Carlos Riquelme, Joan Puigcerver, Rodolphe Jenatton, and Neil Houlsby.
\newblock Multimodal contrastive learning with limoe: the language-image mixture of experts.
\newblock {\em Advances in Neural Information Processing Systems}, 35:9564--9576, 2022.

\bibitem{more}
Han Zhao, Wenxuan Song, Donglin Wang, Xinyang Tong, Pengxiang Ding, Xuelian Cheng, and Zongyuan Ge.
\newblock More: Unlocking scalability in reinforcement learning for quadruped vision-language-action models.
\newblock {\em arXiv preprint arXiv:2503.08007}, 2025.

\bibitem{rectifiedflow}
Qiang Liu.
\newblock Rectified flow: A marginal preserving approach to optimal transport.
\newblock {\em arXiv preprint arXiv:2209.14577}, 2022.

\bibitem{flowmatching}
Yaron Lipman, Ricky~TQ Chen, Heli Ben-Hamu, Maximilian Nickel, and Matt Le.
\newblock Flow matching for generative modeling.
\newblock {\em arXiv preprint arXiv:2210.02747}, 2022.

\bibitem{siglip}
Xiaohua Zhai, Basil Mustafa, Alexander Kolesnikov, and Lucas Beyer.
\newblock Sigmoid loss for language image pre-training.
\newblock In {\em Proceedings of the IEEE/CVF international conference on computer vision}, pages 11975--11986, 2023.

\bibitem{moemultitask}
Jiaqi Ma, Zhe Zhao, Xinyang Yi, Jilin Chen, Lichan Hong, and Ed~H Chi.
\newblock Modeling task relationships in multi-task learning with multi-gate mixture-of-experts.
\newblock In {\em Proceedings of the 24th ACM SIGKDD international conference on knowledge discovery \& data mining}, pages 1930--1939, 2018.

\end{thebibliography}
}

%%%%%%%%%%%%%%%%%%%%%%%%%%%%%%%%%%%%%%%%%%%%%%%%%%%%%%%%%%%%

\clearpage

\appendix
\appendix
\section{Data Collection System} \label{Sec: appendixData}
\vspace{-8mm}

\begin{figure}[ht]
    \centering
  \includegraphics[width=0.85\textwidth]{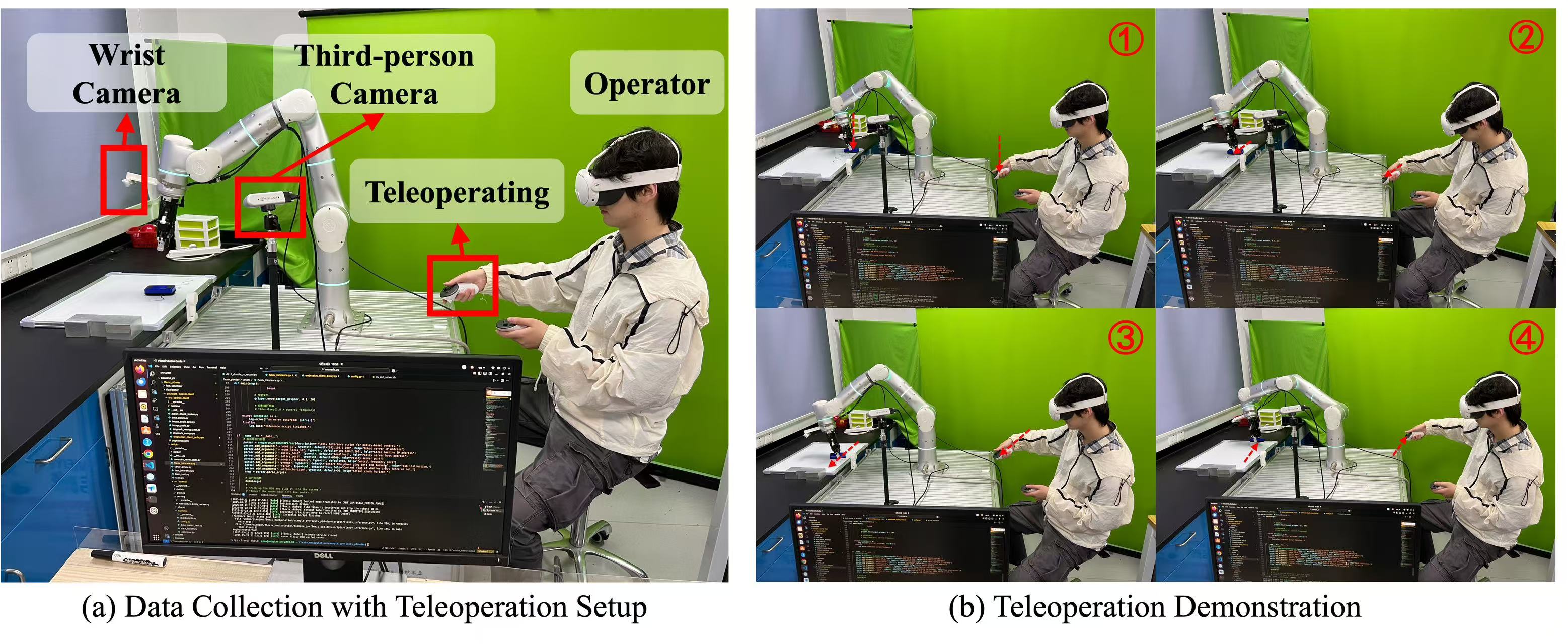}
  \vspace{-2mm}
  \caption{Our data collection system setup.}
  \label{Fig: appen_data_collect}
\end{figure}
\vspace{-4mm}

Our data collection system is depicted in Figure~\ref{Fig: appen_data_collect}(a). The setup features a robotic arm equipped with both a wrist-mounted camera and a static third-person view camera to capture diverse visual perspectives. An operator, wearing a Quest 3 headset and using hand-held controllers, teleoperates the robot arm. A nearby computer runs the necessary software for data acquisition, which includes programs for interfacing with the robot, synchronizing sensor streams, and managing communication with the VR teleoperation hardware. Figure~\ref{Fig: appen_data_collect}(b) illustrates the operator teleoperating the robotic arm to collect demonstration data for the \textit{wipe board} task.

\vspace{-4mm}
\section{Training Details} \label{Sec: appendix1}
\vspace{-4mm}

The models were mainly trained on compute nodes equipped with $8 \times$ \textsc{NVIDIA} \textsc{RTX} 4090 GPUs (24\,GB VRAM each), 64 physical CPU cores, and 251\,GB system RAM, using Adam optimization ($\beta_1=0.9$, $\beta_2=0.95$) with a peak learning rate of $2.5\times10^{-5}$ decaying to $2.5\times10^{-6}$ over 30,000 steps. Multi-task training utilized data parallelism across 2 GPUs (global batch size 16, effective 2048 via gradient accumulation) as additional GPUs provided diminishing returns due to communication overhead, completing 30,000 steps in $\sim$12 hours, while single-task training used 1 GPU for 10,000 steps ($\sim$9 hours), both employing \texttt{bfloat16} precision with gradient clipping ($\|\nabla\|=1.0$).

The dimensionalities and key parameters of ForceVLA's core processing modules: Input Projections, the FVLMoE block, and the Action Output Head are detailed in Table~\ref{tab:model_architecture_focused}.

\begin{table}[htbp]
  \caption{Focused view of ForceVLA's key architectural components: Input Projections, FVLMoE, and Action Output. Dimensions are indicative (e.g., $D_{\text{VLM}}$, $D_{\text{act\_e}}$ for VLM and Action Expert).}
  \label{tab:model_architecture_focused}
  \centering
  \small % Use smaller font for the table
  \renewcommand{\theadfont}{\small\bfseries} % Header font
  \renewcommand{\theadalign}{ll} % Align headers left
  \setlength{\tabcolsep}{4pt} % Adjust column separation
  % Column 1: Sub-Component (left-aligned)
  % Column 2: Key Parameters / Dimensions (left-aligned, wrapping)
  \begin{tabular}{@{} l L{10cm} @{}}
    \toprule
    \thead{Layer} & \thead{Key Parameters / Dimensions} \\
    \midrule
    \multicolumn{2}{@{}l@{}}{\textbf{Input Projections}} \\
    \midrule
    Force Projection             & Linear; Input: 6 (F/T), Output: $D_{\text{VLM}}=2048$ \\
    State Projection             & Linear; Input: $D_{\text{state}}=32$, Output: $D_{\text{act\_e}}=1024$ \\
    Action Projection            & Linear; Input: $D_{\text{action}}=32$, Output: $D_{\text{act\_e}}=1024$ \\
    Action-Time MLP             & 2-layer MLP; Input: $2 \times D_{\text{act\_e}}$, Hidden/Output: $D_{\text{act\_e}}$; Swish activation \\
    \midrule
    \multicolumn{2}{@{}l@{}}{\textbf{FVLMoE Module}} \\
    \midrule
    Input                       & Concatenation: $N_{\text{VL}} \times D_{\text{VLM}}$ (V-L features) \& $1 \times D_{\text{VLM}}$ (Force token) \\
    Pre-MoE Encoder             & Transformer Encoder Block; $D_{\text{model}}=2048, N_{\text{H}}=8, D_{\text{h}}=256$; MLP (expansion factor 1) \\
    MoE Layer                   & Sparse MoE; $E=4$ experts (MLPs: $D_{\text{model}} \to D_{\text{model}}$), Top-$k=1$; Router: $D_{\text{model}} \to E$ \\
    Output Projection           & Linear; Input: $D_{\text{model}}=2048$, Output: $D_{\text{act\_e}}=1024$ \\
    \midrule
    \multicolumn{2}{@{}l@{}}{\textbf{Action Output Head}} \\
    \midrule
    Action Output Projection    & Linear; Input: $D_{\text{act\_e}}=1024$, Output: $D_{\text{action}}=32$ \\
    \bottomrule
  \end{tabular}
\end{table}

% \vspace{-10mm}
\vspace{-4mm}
\section{Router Analysis} \label{Sec: appendix2}
\vspace{-4mm}

\begin{figure}[H]
% \vspace{-15mm}
    \centering
    \includegraphics[width=0.97\textwidth]{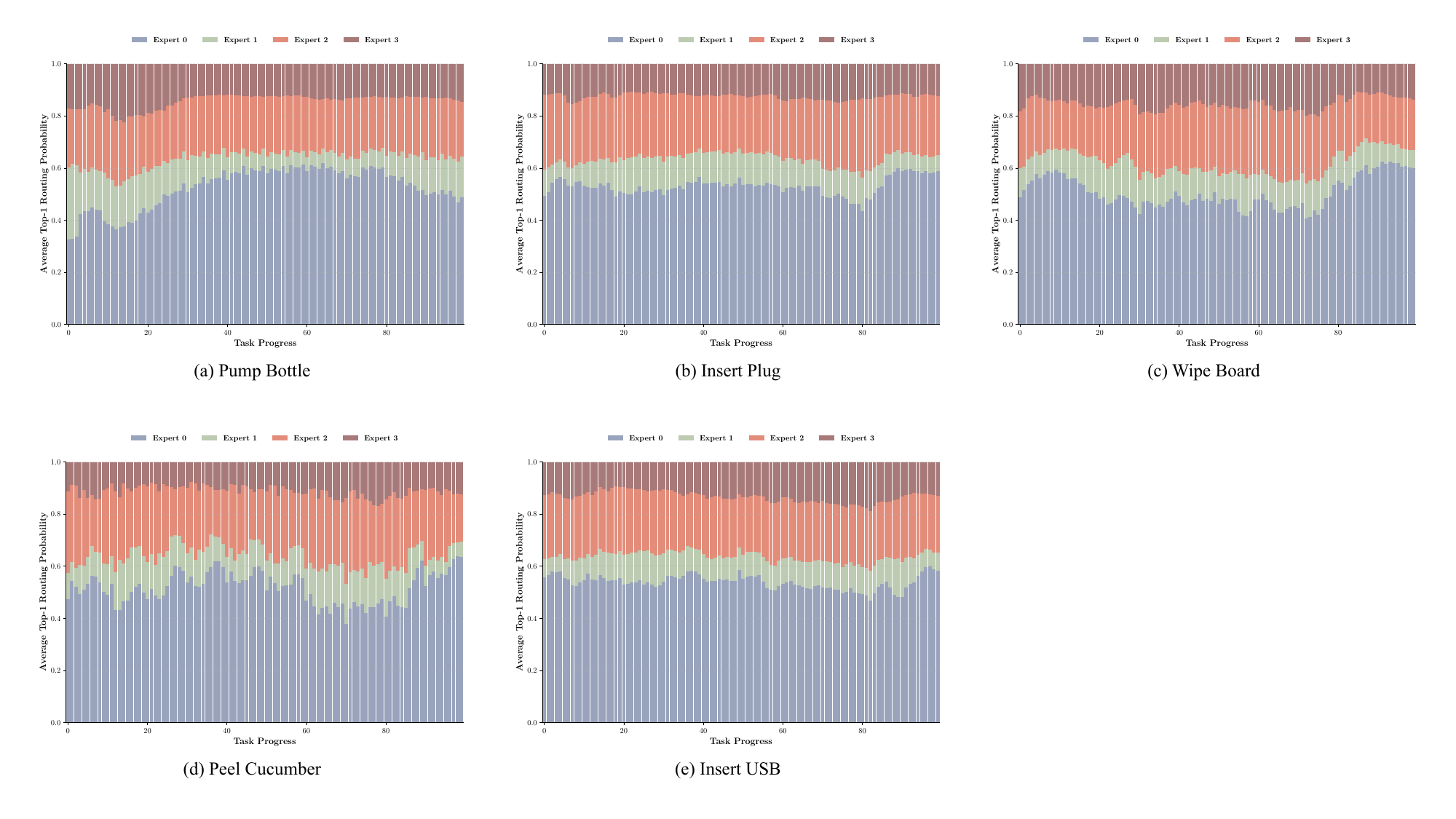}
\vspace{-4mm}
    \caption{Open-loop evaluation of expert load across different task completion percentages for various tasks: (a) Pump Bottle, (b) Insert Plug, (c) Wipe Board, (d) Peel Cucumber, and (e) Insert USB. Each subplot represents the average expert load (vertical axis) as a function of the task completion percentage (horizontal axis) over the episodes in the test dataset.}
    \label{Fig: ForceVLA_router}
\end{figure}
\vspace{-4mm}

To analyze routing dynamics, we first measured the probability distribution over expert selections for each token as it was processed by the router in the MoE module. For variable-length episodes, we applied percentile-based normalization: each task ($\sim$10 episodes) was processed by segmenting every episode’s token sequence into 100 temporally equidistant intervals, computing the mean top-1 probability per segment, and then averaging these means across episodes. This ensured cross-episode comparability while preserving temporal routing dynamics.

As shown in Figure~\ref{Fig: ForceVLA_router}, different tasks exhibit distinct expert utilization patterns. Some tasks (e.g., \textit{insert plug}, \textit{peel cucumber}) show clear temporal specialization, where certain experts dominate specific phases of the task. Others (e.g., \textit{wipe board}) demonstrate a more consistent preference for a single expert throughout execution. These trends suggest that the router learns to allocate computation dynamically across experts based on task-specific semantics and temporal structure. 

What’s more, we found that Expert 0 dominates nearly half of the tokens across multiple tasks. This persistent activation suggests that Expert 0 may function as a general-purpose expert, responsible for the fusion of multiple modalities or routine control primitives that are shared across tasks. Its broad involvement contrasts with the more selective, phase-specific activation of Expert 1 or Expert 3, reinforcing the hypothesis of functional specialization among experts. Such asymmetry in routing frequency reflects not only temporal semantic variance within tasks but also architectural bias toward certain experts, potentially shaped during training.

\vspace{-4mm}
\section{Multi-task Evaluation} \label{Sec: appendix3}
\vspace{-4mm}

\begin{table}[H] % Position specifier (here, top, bottom, page)
  \centering
  \caption{Multi-task joint training success rates (\%). ForceVLA (Ours) demonstrates superior average performance and excels or matches the best performance in all individual tasks. Best performance(s) in each column are in \textbf{bold}.}
  \label{tab:multitask_joint_training_results}
  \renewcommand{\theadfont}{\small\bfseries} % Makes headers small and bold
  \renewcommand{\theadalign}{cc} % Center-aligns content within \thead cells
  \setlength{\tabcolsep}{5pt} % Adjust space between columns (default is 6pt)
  \small % Makes the font size of the table content smaller
  \begin{tabular}{lrrrrr} % l: left-align (Model), r: right-align (for numbers)
    \toprule
    \thead{Model} & \thead{Pump\\Bottle} & \thead{Insert\\Plug} & \thead{Insert\\USB} & \thead{Wipe\\Board} & \thead{Average\\SR} \\
    \midrule
    $\pi_0$-fast w/o F    & 0.0\%           & 0.0\%           & 0.0\%          & 0.0\%          & 0.0\% \\
    $\pi_0$-fast w/ F     & 0.0\%           & 0.0\%           & 0.0\%          & 0.0\%          & 0.0\% \\
    $\pi_0$-base w/o F    & 20.0\%          & 0.0\%           & 0.0\%          & 0.0\%          & 5.0\% \\
    $\pi_0$-base w/ F     & 50.0\%          & \textbf{100.0\%} & \textbf{10.0\%} & 10.0\%         & 42.5\% \\
    \textbf{ForceVLA (Ours)} & \textbf{80.0\%} & \textbf{100.0\%} & \textbf{10.0\%} & \textbf{80.0\%} & \textbf{67.5\%} \\
    \bottomrule
  \end{tabular}
\end{table}
\vspace{-4mm}

The results of joint multi-task learning are detailed in Table~\ref{tab:multitask_joint_training_results}. Notably, both $\pi_0$-fast variants (0\% average success rate) failed to acquire skills in this setting, indicating their limited capacity for diverse, concurrent learning. The $\pi_0$-base w/o F model also performed poorly (5\% average success rate), managing only 20\% success on a single task. Adding direct force input ($\pi_0$-base w/ F) improved the average performance to 42.5\%, primarily due to its success in the \textit{insert plug} task. Our \textbf{ForceVLA} model demonstrated superior multi-task capabilities, achieving the highest average success rate of 67.5\%. It obtained 80\% success in both \textit{pump bottle} and \textit{wipe board} tasks, and matched top performance in \textit{insert plug} (100\%) and \textit{insert usb} (10\%). This robust performance across multiple distinct tasks indicates ForceVLA's capacity for concurrent skill learning, proficient instruction following for varied goals, and highlights the role of its FVLMoE architecture in utilizing multimodal cues, particularly force, within a shared policy.

\section{Real-world Experiments Visualization} \label{Sec: appendix4}

In this section, we present key frames from real-world experiment videos. Each visualization contrasts failure cases of baseline models with successful task completions by our ForceVLA model under similar conditions.

\begin{figure}[ht]
    \centering
    \includegraphics[width=\textwidth]{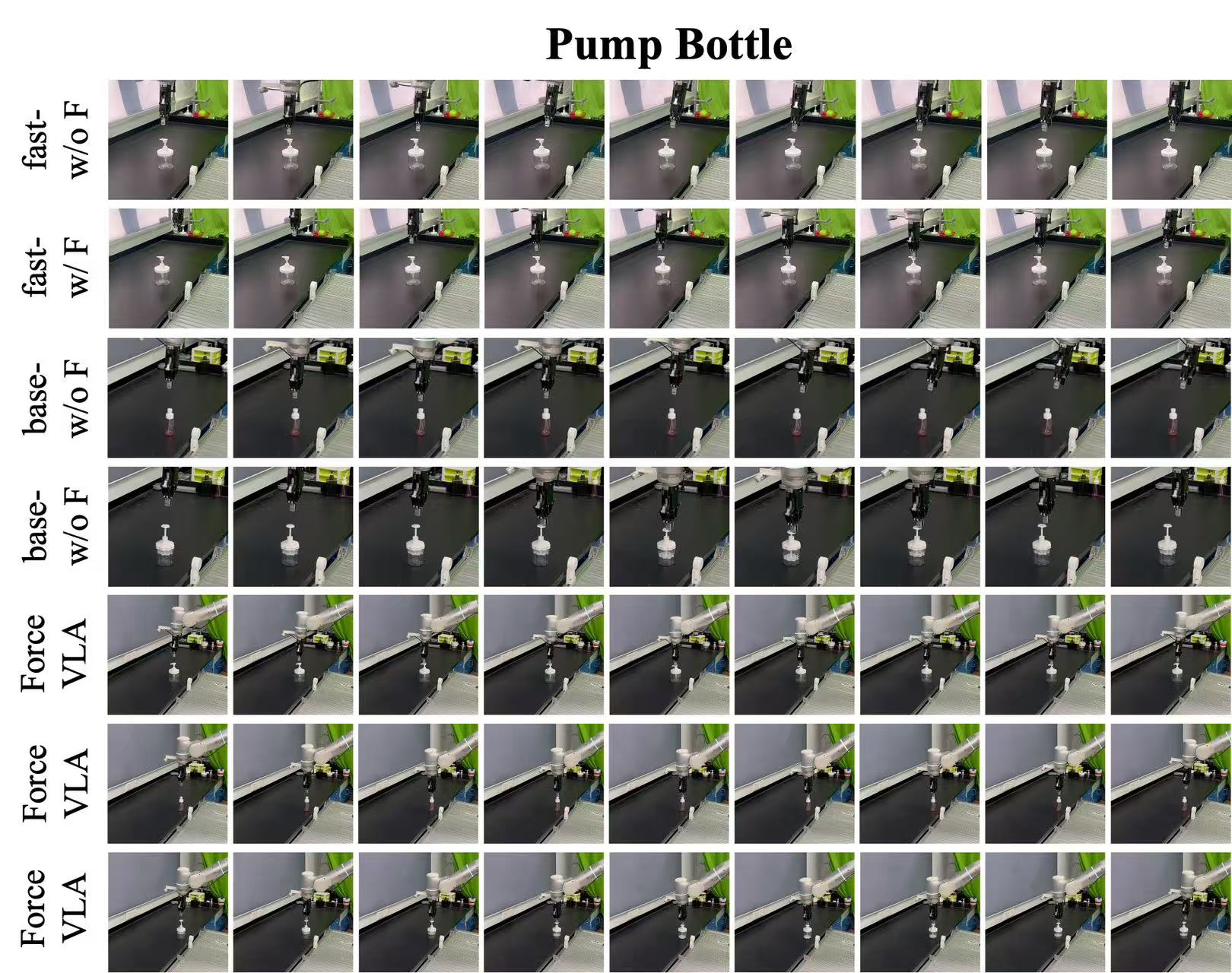}
    \caption{Key frames from Pump Bottle task videos.}
    \label{Fig: appen_pump_bottle}
\end{figure}

\begin{figure}[ht]
    \centering
  \includegraphics[width=\textwidth]{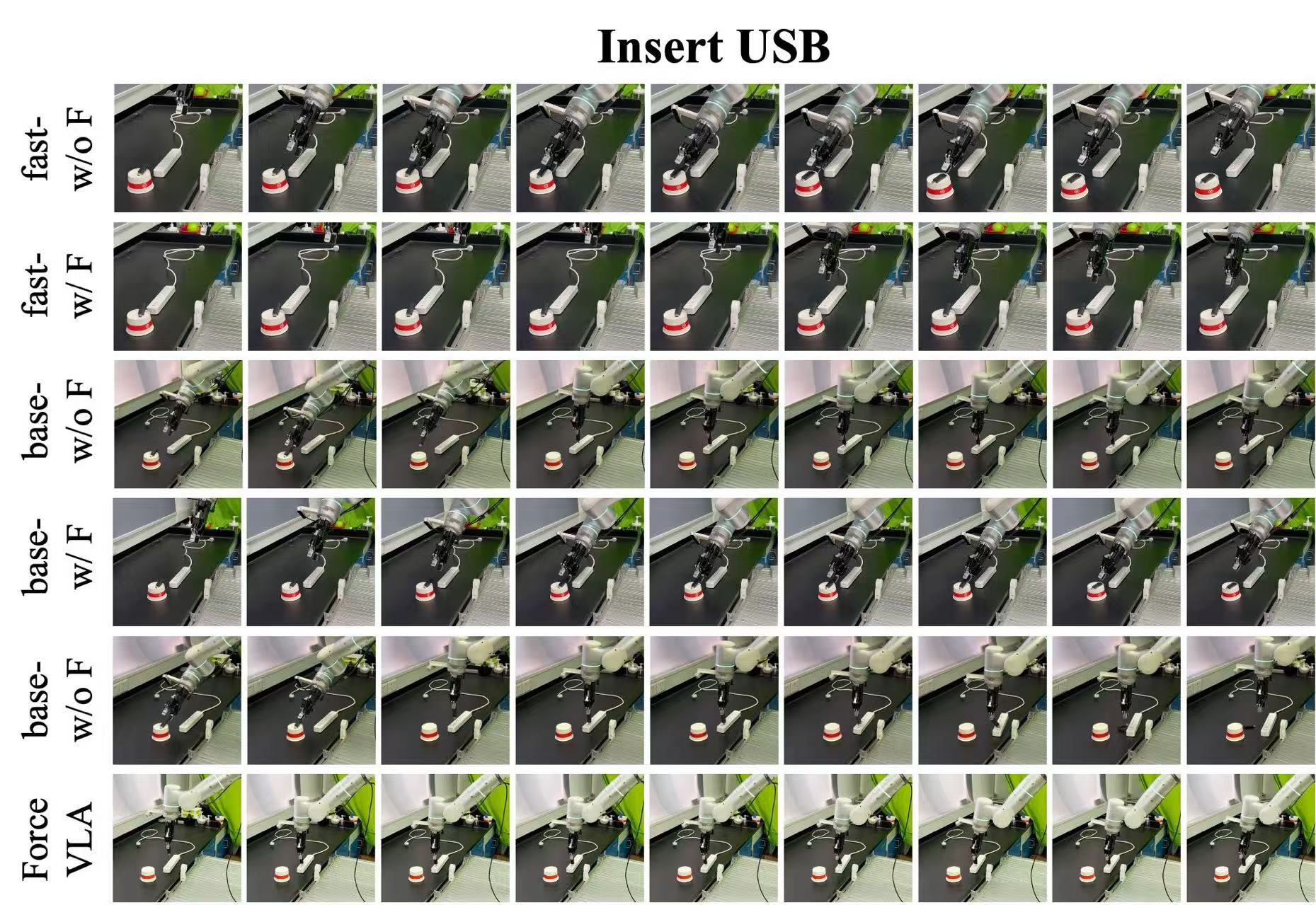}
  \caption{Key frames from Insert USB task videos.}
  \label{Fig: appen_insert_usb}
\end{figure}

\begin{figure}[ht]
    \centering
  \includegraphics[width=\textwidth]{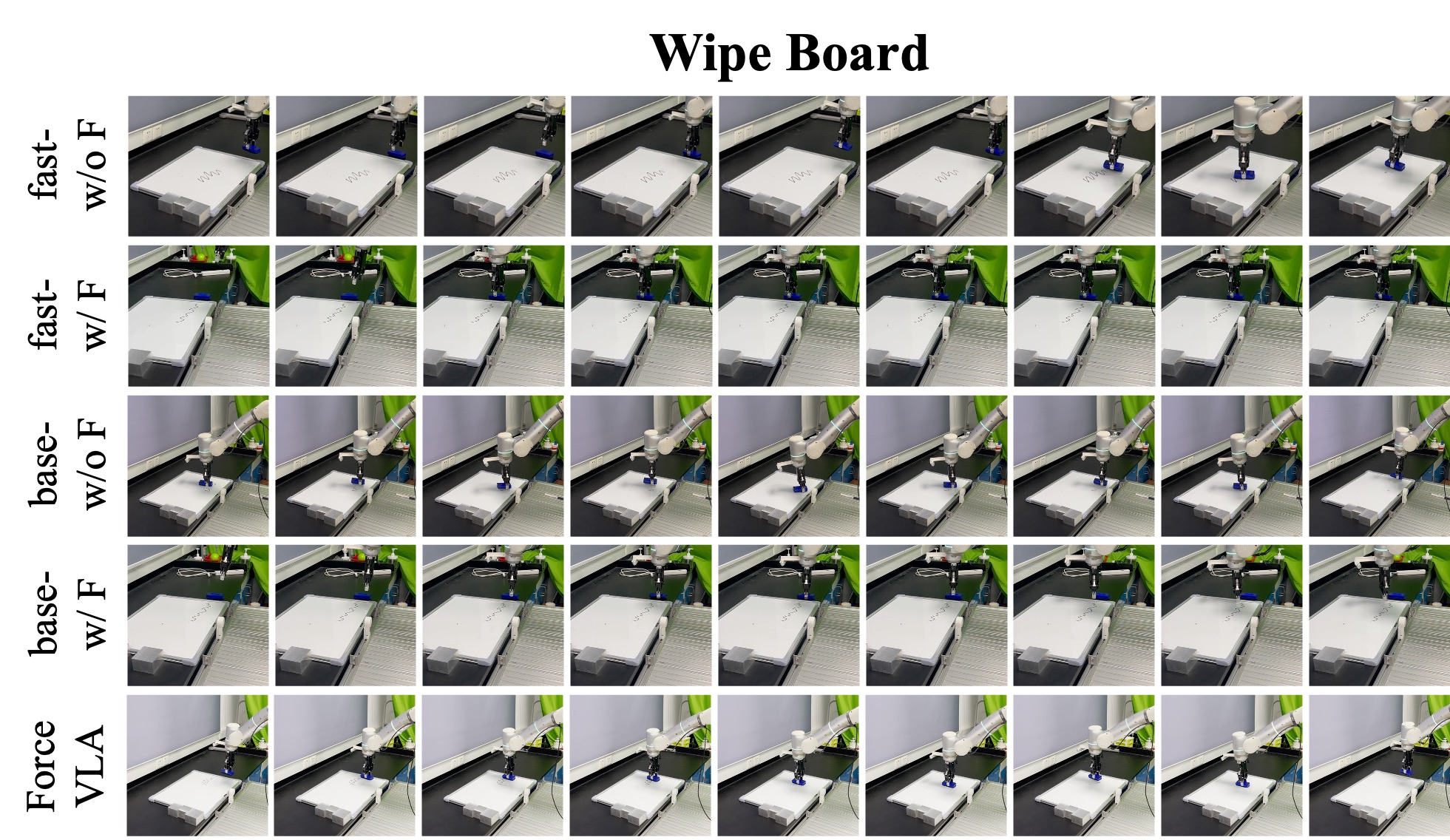}
  \caption{Key frames from Wipe Board task videos.}
  \label{Fig: appen_wipe_board}
\end{figure}

\begin{figure}[ht]
    \centering
  \includegraphics[width=\textwidth]{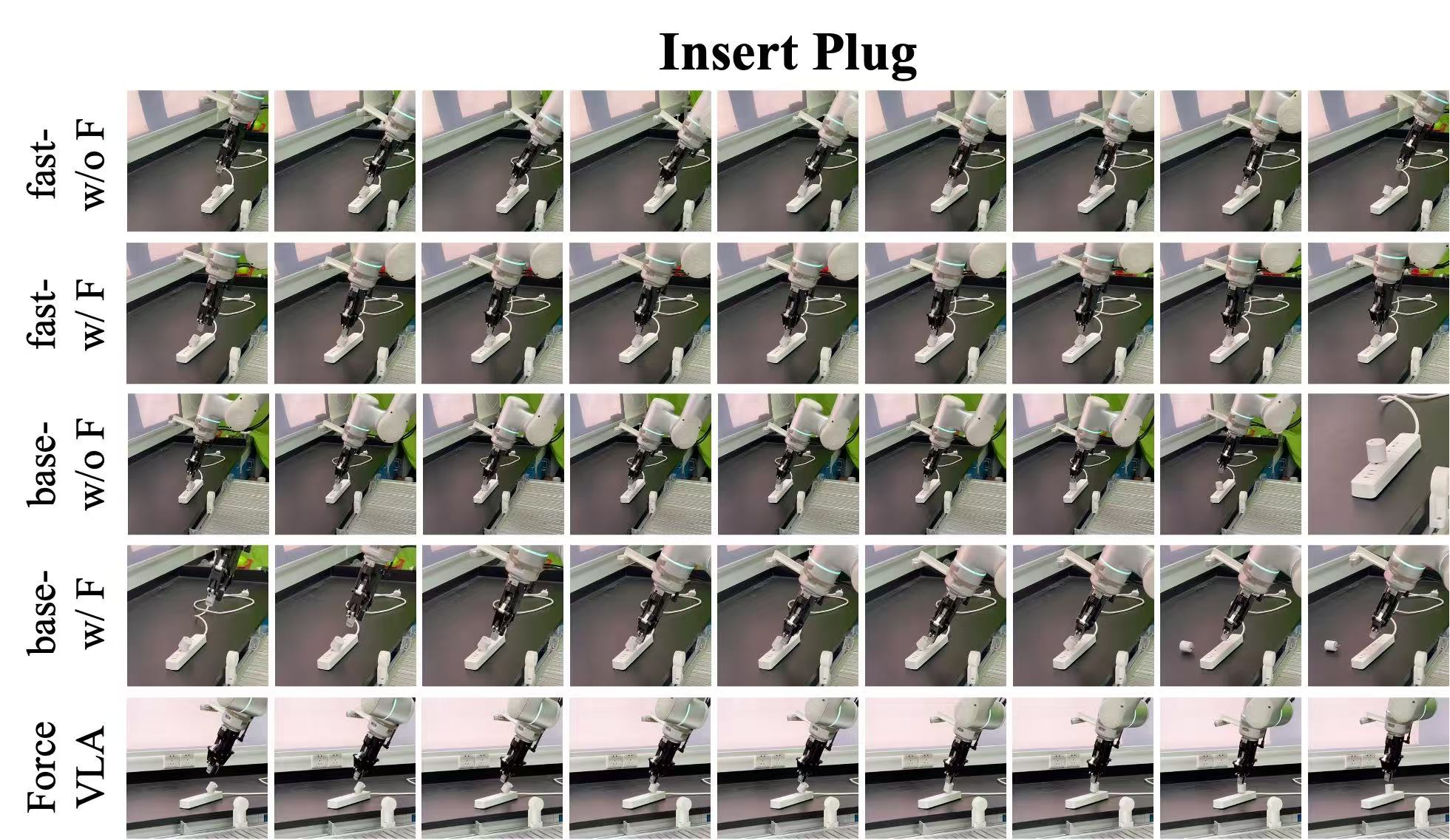}
  \caption{Key frames from Insert Plug task videos.}
  \label{Fig: appen_plug}
\end{figure}

\begin{figure}[ht]
    \centering
  \includegraphics[width=\textwidth]{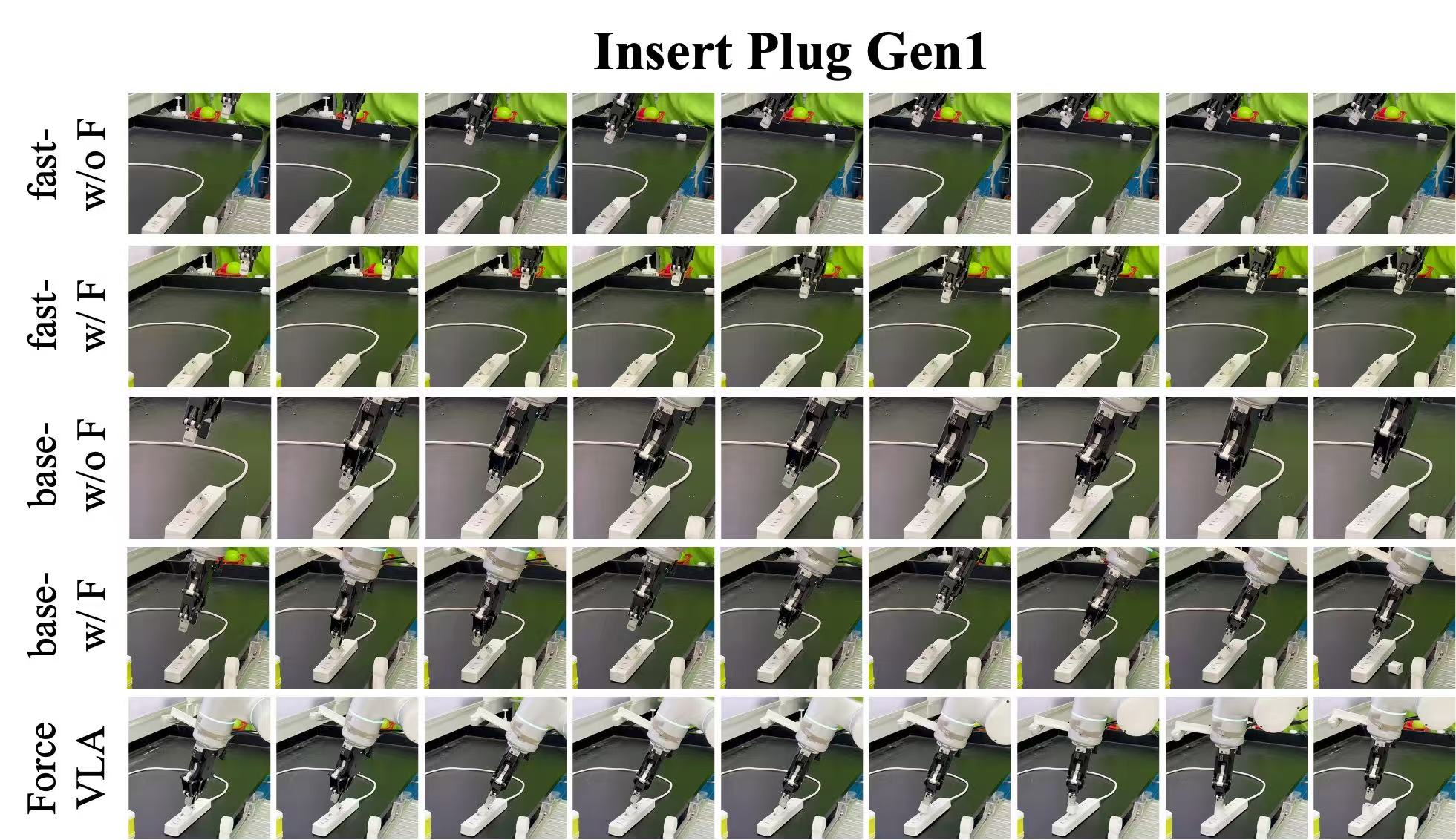}
  \caption{Key frames from Insert Plug Generalization task 1 videos.}
  \label{Fig: appen_plug_gen1}
\end{figure}

\begin{figure}[ht]
    \centering
  \includegraphics[width=\textwidth]{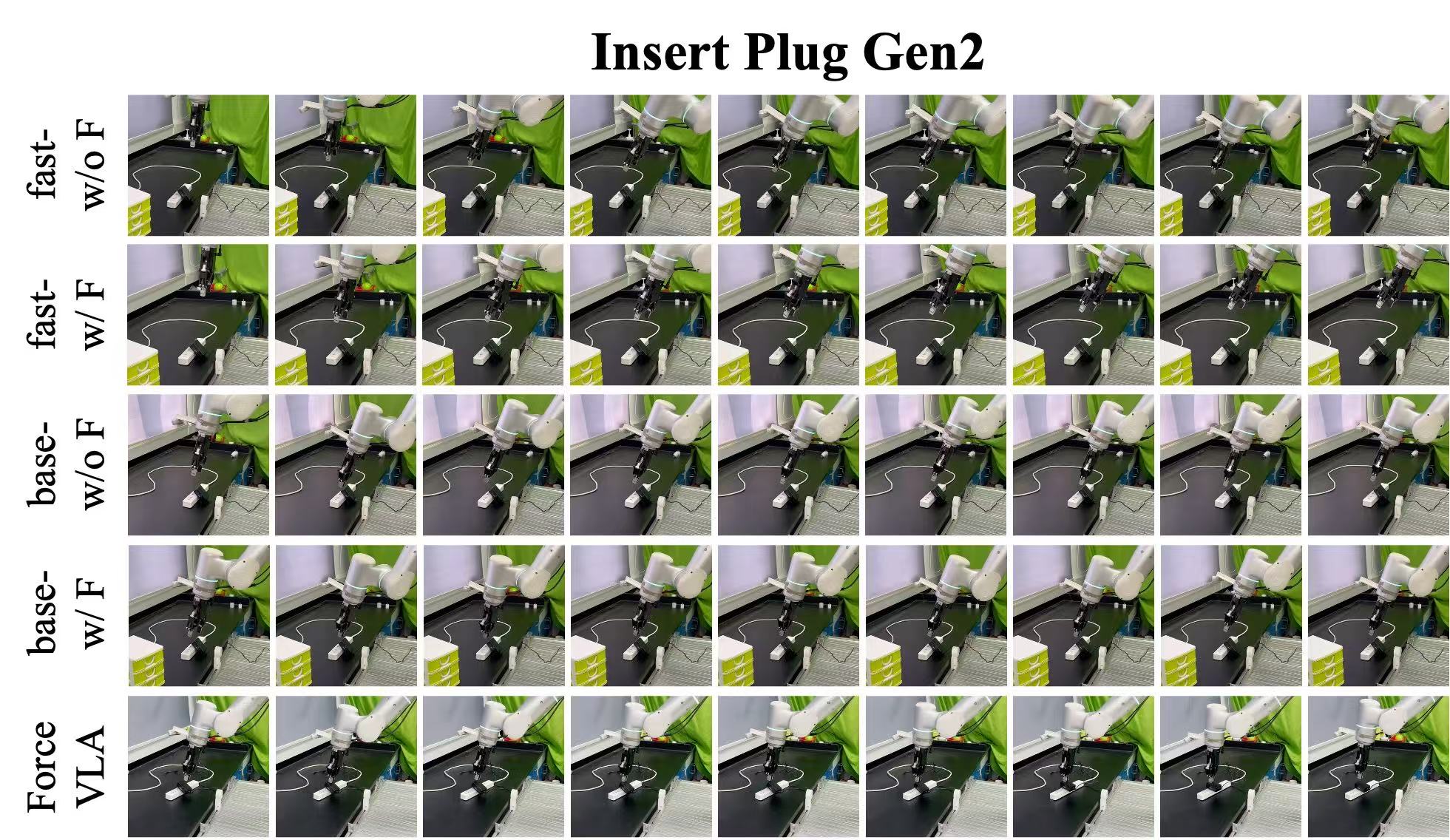}
  \caption{Key frames from Insert Plug Generalization task 2 videos.}
  \label{Fig: appen_plug_gen2}
\end{figure}

\begin{figure}[ht]
    \centering
  \includegraphics[width=\textwidth]{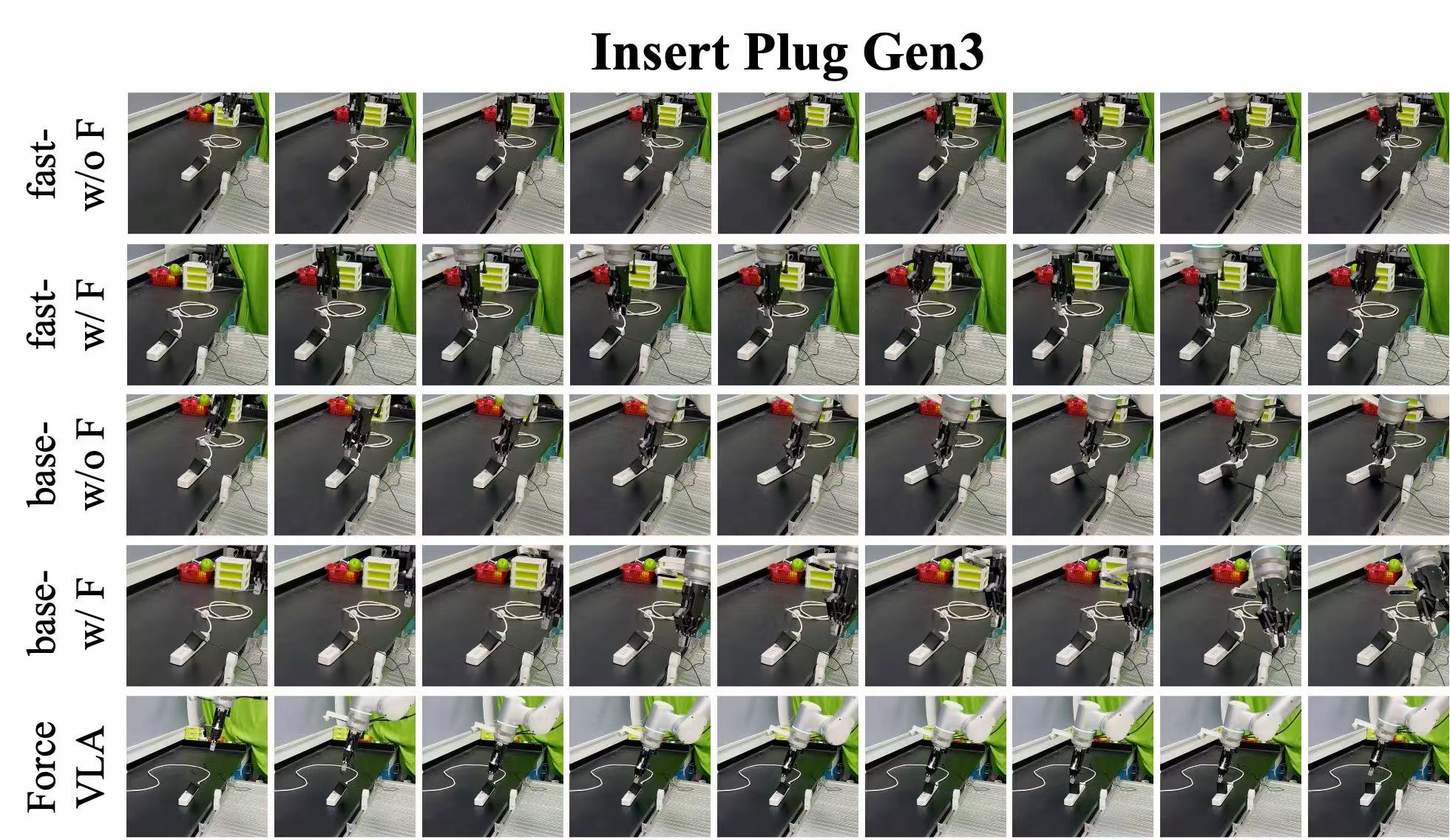}
  \caption{Key frames from Insert Plug Generalization task 3 videos.}
  \label{Fig: appen_plug_gen3}
\end{figure}

\begin{figure}[ht]
    \centering
  \includegraphics[width=\textwidth]{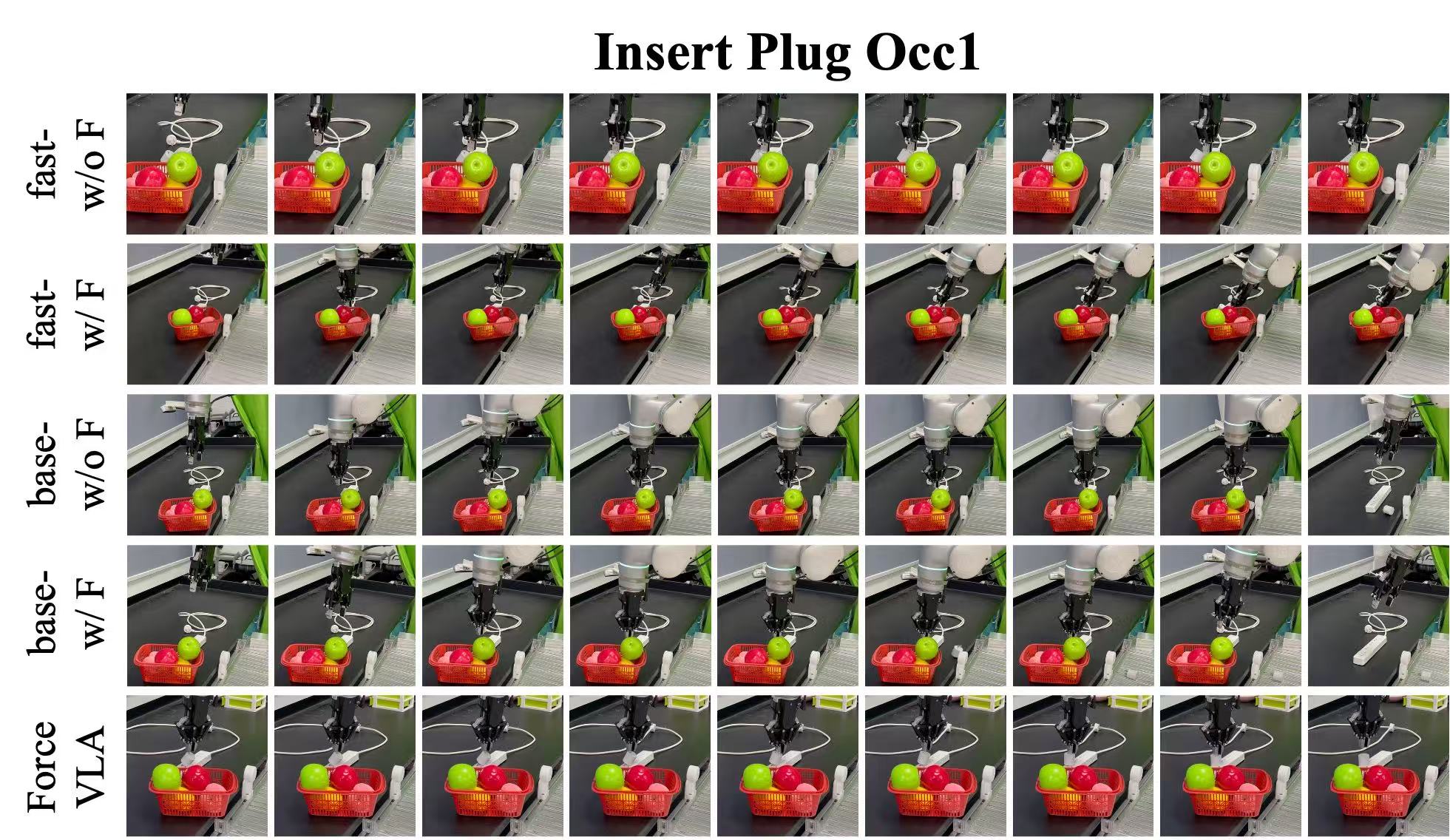}
  \caption{Key frames from Insert Plug Occlusion task 1 videos.}
  \label{Fig: appen_plug_occ1}
\end{figure}

\begin{figure}[ht]
    \centering
  \includegraphics[width=\textwidth]{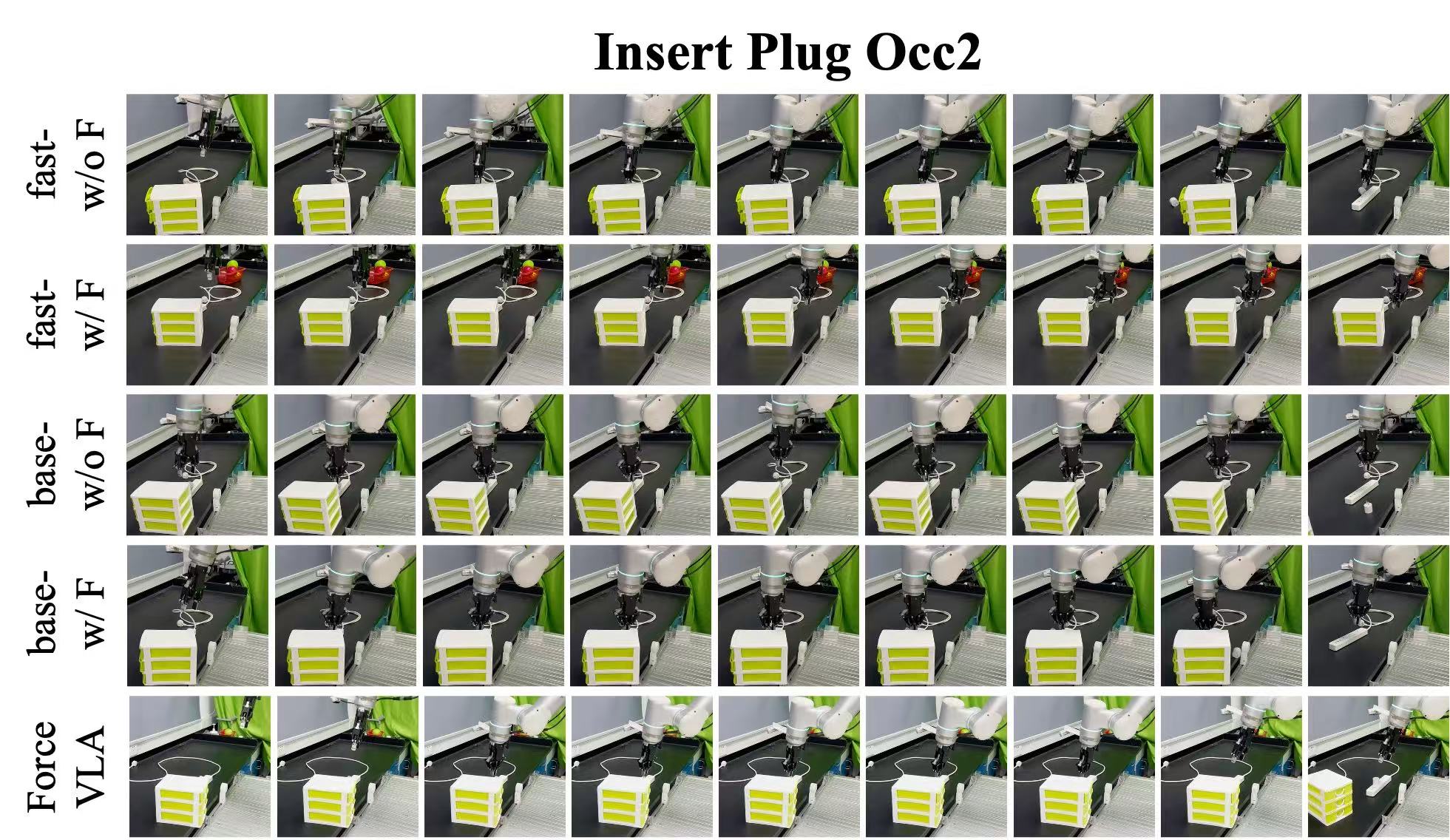}
  \caption{Key frames from Insert Plug Occlusion task 2 videos.}
  \label{Fig: appen_plug_occ2}
\end{figure}

\begin{figure}[ht]
    \centering
  \includegraphics[width=\textwidth]{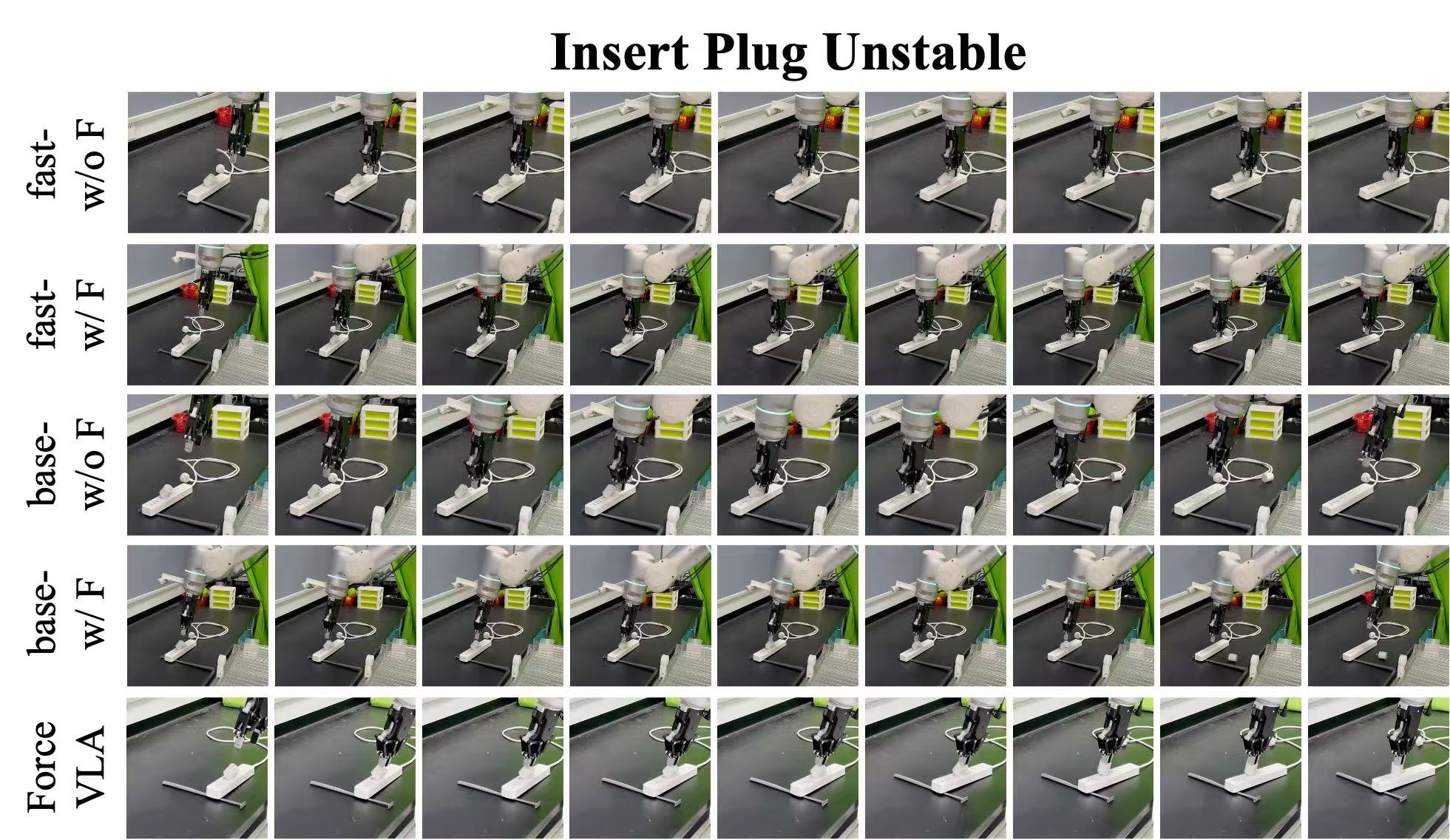}
  \caption{Key frames from Insert Plug Unstable task videos.}
  \label{Fig: appen_plug_unstable}
\end{figure}

% \clearpage

% \input{sections/12_check_list}

%%%%%%%%%%%%%%%%%%%%%%%%%%%%%%%%%%%%%%%%%%%%%%%%%%%%%%%%%%%%

\end{document}